\documentclass{article}

\usepackage[final]{neurips_2022}
\usepackage{wrapfig}
\usepackage{subcaption}

\usepackage{microtype}
\usepackage{graphicx}
\usepackage{booktabs} 
\usepackage{xcolor}
\usepackage{bm,bbm}
\usepackage{algorithm}
\usepackage{caption}
\usepackage[utf8]{inputenc} 
\usepackage[T1]{fontenc}    
\usepackage{hyperref}       
\usepackage{url}            
\usepackage{booktabs}       
\usepackage{amsfonts}       
\usepackage{nicefrac}       
\usepackage{microtype}      
\usepackage{xcolor}  
\usepackage{amsmath, amsthm, amssymb, multirow, paralist}
\usepackage{mathtools}
\usepackage{bm,bbm}
\usepackage{hyperref}
\hypersetup{colorlinks=true,breaklinks=true}

\usepackage{algorithm}
\usepackage{listings}

\newtheorem{theorem}{Theorem}
\newtheorem*{theorem*}{Theorem}

\def \R {\mathbb{R}}

\title{FiLM: Frequency improved Legendre Memory Model for Long-term
Time Series Forecasting}

\author{%
Tian Zhou\thanks{equal contribution}\quad Ziqing Ma$^*$ \quad  Xue Wang\quad Qingsong Wen\quad  Liang Sun \\ \textbf{Tao Yao}\quad \textbf{Wotao Yin} \quad  \textbf{Rong Jin} \\
\texttt{\{tian.zt,maziqing.mzq,xue.w,qingsong.wen,liang.sun\}@alibaba-inc.com }\\
\texttt{\{tao.yao,wotao.yin,jinrong.jr\}@libaba-inc.com}\\
}

\begin{document}

\maketitle

\begin{abstract}
  
  Recent studies have shown that deep learning models such as RNNs and Transformers have brought significant performance gains for long-term forecasting of time series because they effectively utilize historical information. We found, however, that there is still great room for improvement in how to preserve historical information in neural networks while avoiding overfitting to noise presented in the history. Addressing this allows better utilization of the capabilities of deep learning models. To this end, we design a \textbf{F}requency \textbf{i}mproved \textbf{L}egendre \textbf{M}emory model, or {\bf FiLM}: it applies Legendre Polynomials projections to approximate historical information, uses Fourier projection to remove noise, and adds a low-rank approximation to speed up computation. Our empirical studies show that the proposed FiLM significantly improves the accuracy of state-of-the-art models in multivariate and univariate long-term forecasting by (\textbf{20.3\%}, \textbf{22.6\%}), respectively. We also demonstrate that the representation module developed in this work can be used as a general plug-in to improve the long-term prediction performance of other deep learning modules. Code is available at https://github.com/tianzhou2011/FiLM/.

\end{abstract}

\section{Introduction}\label{sec_intro}
\begin{wrapfigure}[12]{r}{0.7\textwidth}
\vskip -0.12in
\centering
\includegraphics[width=0.49\linewidth]{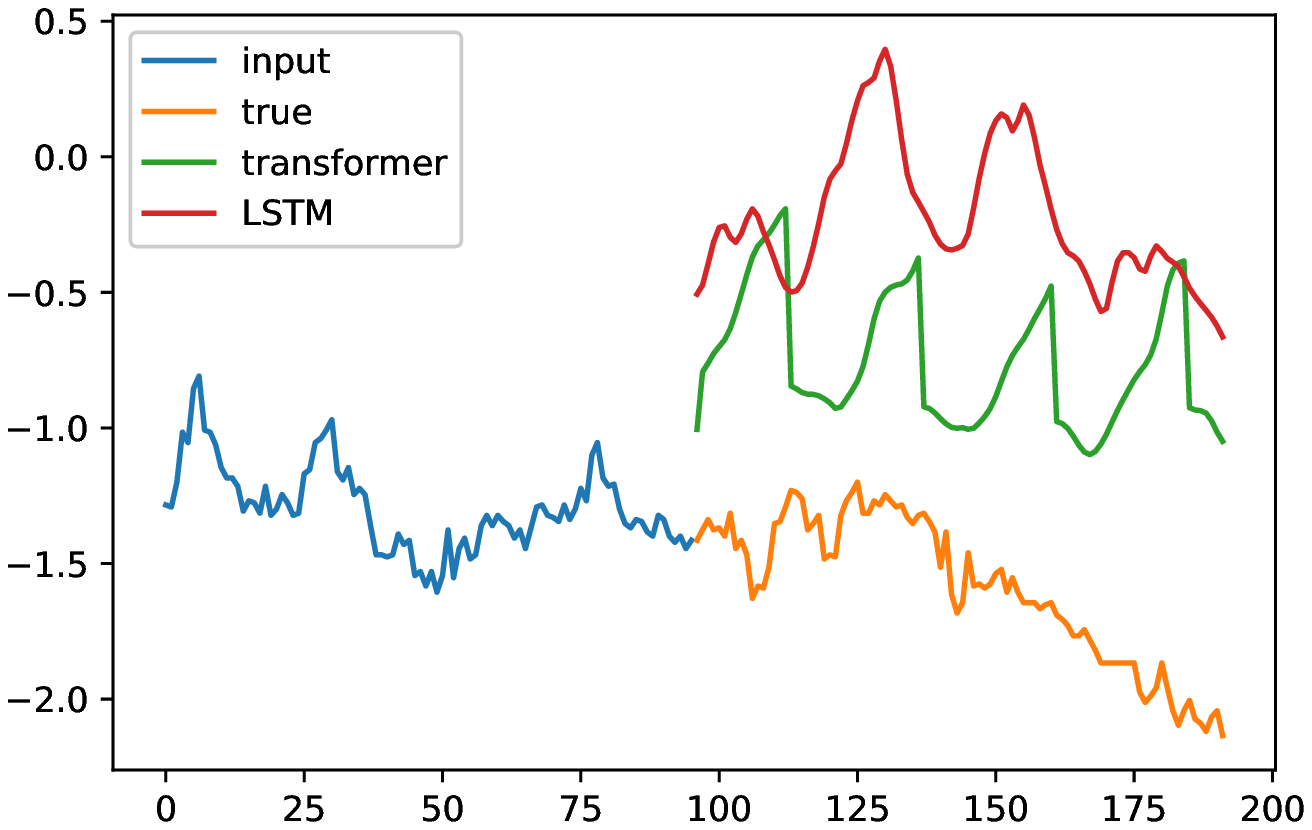}
\includegraphics[width=0.49\linewidth]{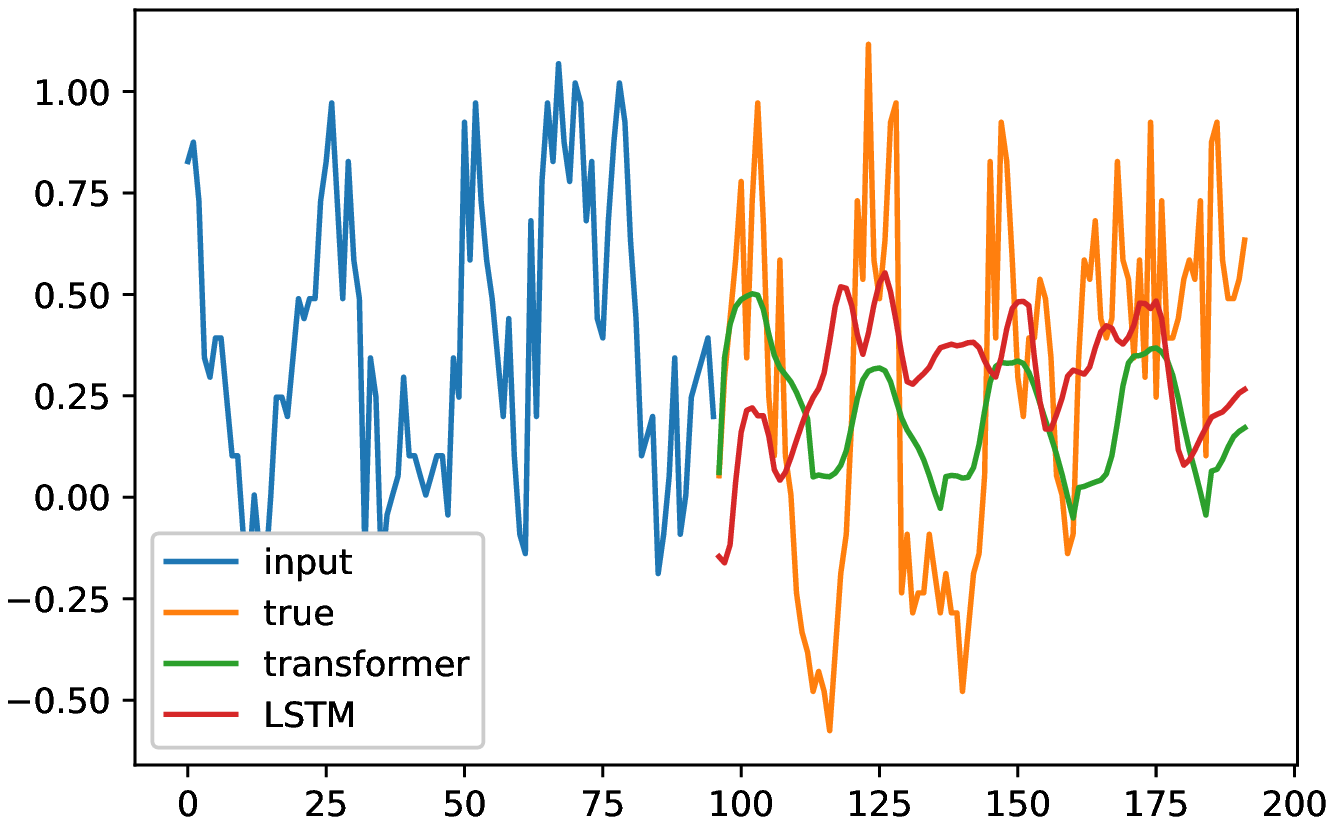}
\vskip -0.10in
\caption{The discrepancy between ground truth and forecasting output from vanilla Transformer and LSTM on a real-world ETTh1 dataset Left: trend shift. Right: seasonal shift.}
\label{fig:descrepancy}
\vskip -0.20in
\end{wrapfigure}

Long-term forecasting refers to making predictions based on the history for a long horizon in the future, as opposed to short-term forecasting. Long-term time series forecasting has many key applications in energy, weather, economics, transportation, and so on. It is more challenging than ordinary time series forecasting.
Some of the challenges in long-term forecasting include long-term time dependencies, susceptibility to error propagation, complex patterns, and nonlinear dynamics. These challenges make accurate predictions generally impossible for traditional learning methods such as ARIMA.
Although RNN-like deep learning methods have made breakthroughs in time series forecasting~\cite{deep-state-space-models-for-time-series-forecasting,DBLP:journals/corr/FlunkertSG17-deepAR}, they often suffer from problems such as gradient vanishing/exploding ~\cite{DBLP:icml/On-the-difficult-gradient-vanishing-explode}, which limits their practical performance. Following the success of Transformer~\cite{attention_is_all_you_need} in the NLP and CV communities~\cite{attention_is_all_you_need,Bert/NAACL/Jacob,Transformers-for-image-at-scale/iclr/DosovitskiyB0WZ21,DBLP:Global-filter-FNO-in-cv}, it also shows promising performance in capturing long-term dependencies~\cite{haoyietal-informer-2021, Autoformer, FedFormer, wen2022transformers} for time series forecasting. We provide an overview of this line of work (including deep recurrent networks and efficient Transformers) in the appendix \ref{App:related_works}.
In order to achieve accurate predictions, many deep learning researchers increase the complexity of their models, hoping that they can capture critical and intricate historical information. These methods, however, cannot achieve it. Figure ~\ref{fig:descrepancy} compares the ground truth time series of the real-world ETTm1 dataset with the predictions of the vanilla Transformer method and the LSTM model~\cite{attention_is_all_you_need,hochreiter1997long}~ \cite{ haoyietal-informer-2021}. It is observed that the prediction is completely off from the distribution of the ground truth. We believe that such errors come from these models miscapturing noise while attempting to preserve the true signals. We conclude that two keys in accurate forecasting are: 1) how to capture critical historical information as complete as possible; and 2) how to effectively remove the noise. 
Therefore, to avoid a derailed forecast, we cannot improve a model by simply making it more complex. Instead, we shall consider a robust representation of the time series that can capture its important patterns without noise.

\begin{wrapfigure}[12]{r}{0.7\textwidth}
\vskip -0.22in
\centering
\includegraphics[width=0.49\linewidth]{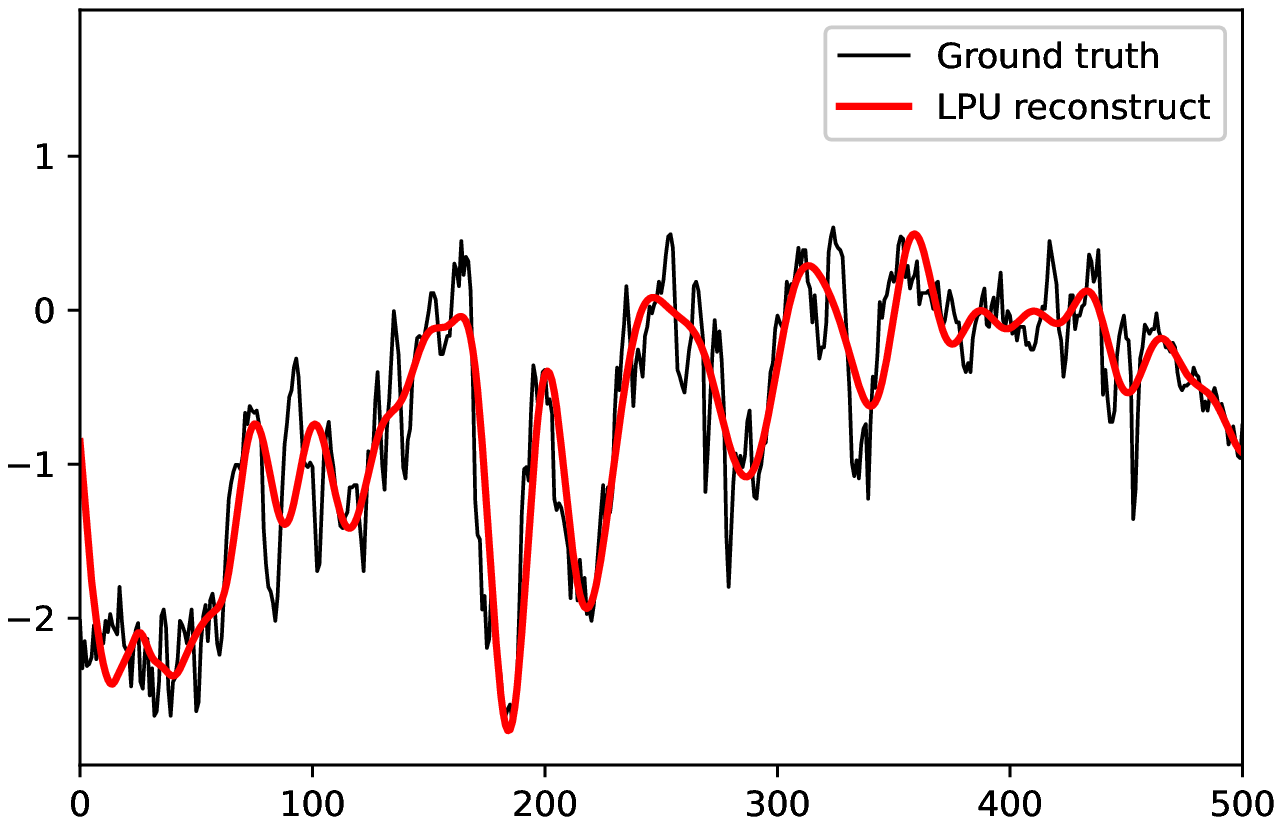}
\includegraphics[width=0.49\linewidth]{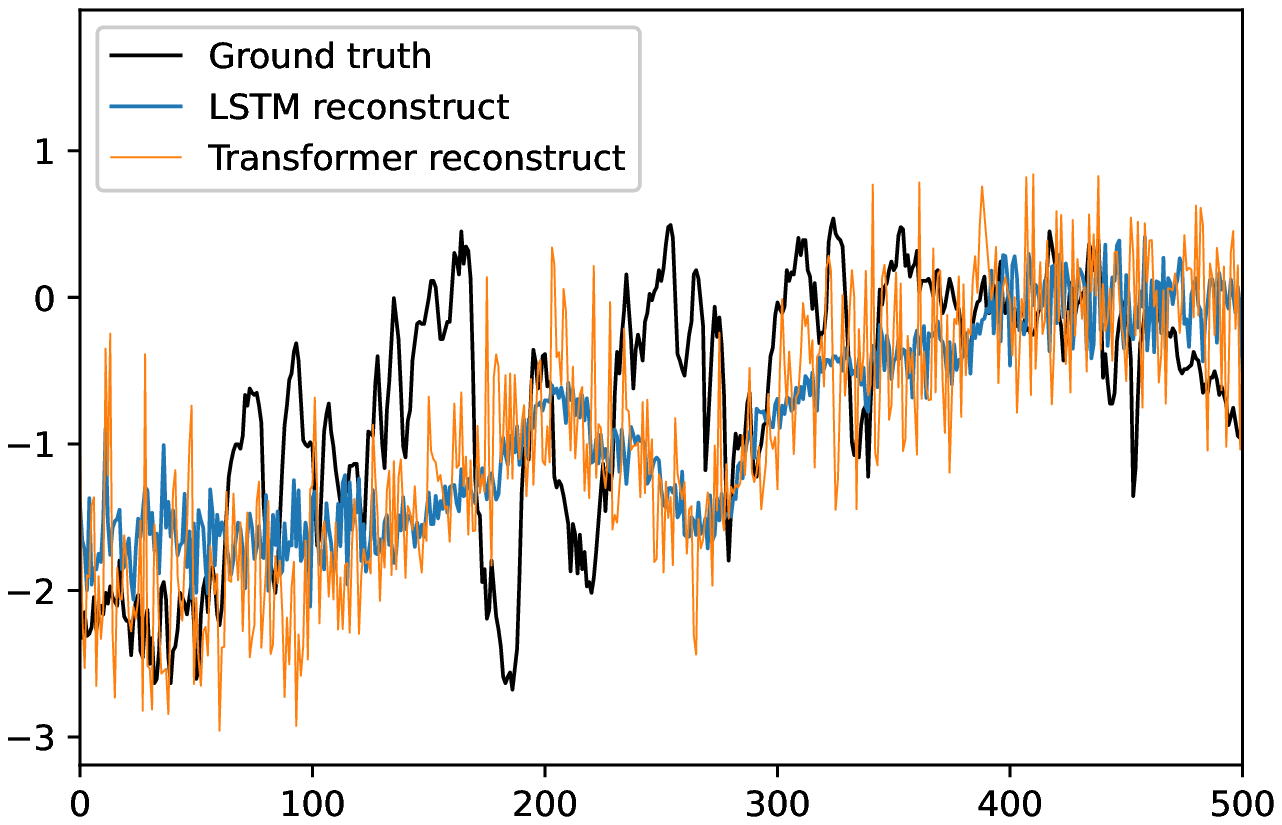}
\vskip -0.1in
\caption{Data recovery with Autoencoder structure: recovery a 1024-length data with a bottleneck of 128 parameters. Left: Legendre Projection Unit. Right: LSTM and vanilla Transformer.}
\label{fig:recover}
\vskip -0.20in
\end{wrapfigure}

This observation motivates us to switch our view from long-term time series forecasting to long-term sequence compression. Recursive memory model~\cite{LMU,S4,LSSL,Hippo} has achieved impressive results in function approximation tasks. \cite{LMU} designed a recursive memory unit (LMU) using Legendre projection, which provides a good representation for long time series. $S4$ model~\cite{S4} comes up with another recursive memory design for data representation and significantly improves state-of-the-art results for Long-range forecasting benchmark (LRA)~\cite{LRA}. However, when coming to long-term time series forecasting, these approaches fall short of the Transformer-based methods' state-of-the-art performance. A careful examination reveals that these data compression methods are powerful in recovering the details of historical data compared to LSTM/Transformer models, as revealed in Figure~\ref{fig:recover}. However, they are vulnerable to noisy signals as they tend to overfit all the spikes in the past, leading to limited long-term forecasting performance. It is worth noting that, Legendre Polynomials employed in LMP~\cite{LMU} is just a special case in the family of Orthogonal Polynomials (OPs). OPs (including Legendre, Laguerre, Chebyshev, etc.) and other orthogonal basis (Fourier and Multiwavelets) are widely studied in numerous fields and recently brought in deep learning~\cite{wang2018multilevel,Multiwavelet-based-Operator-Learning,FedFormer,LMU,Hippo}. A detailed review can be found in Appendix \ref{App:related_works}.



The above observation inspires us to develop methods for accurate and robust representations of time series data for future forecasting, especially long-term forecasting. The proposed method significantly outperforms existing long-term forecasting methods on multiple benchmark datasets by integrating those representations with powerful prediction models. As the first step towards this goal, we directly exploit the Legendre projection, which is used by LMU~\cite{LMU} to update the representation of time series with fixed-size vectors dynamically. This projection layer will then be combined with different deep learning modules to boost forecasting performance. The main challenge with directly using this representation is the dilemma between information preservation and data overfitting, i.e., the larger the number of Legendre projections is, the more the historical data is preserved, but the more likely noisy data will be overfitted. Hence, as a second step, to reduce the impact of noisy signals on the Legendre projection, we introduce a layer of dimension reduction by a combination of Fourier analysis and low-rank matrix approximation. More specifically, we keep a large dimension representation from the Legendre projection to ensure that all the important details of historical data are preserved. We then apply a combination of Fourier analysis and low-rank approximation to keep the part of the representation related to low-frequency Fourier components and the top eigenspace to remove the impact of noises. Thus, we can not only capture the long-term time dependencies, but also effective reduce the noise in long-term forecasting. We refer to the proposed method as \textbf{F}requency \textbf{i}mproved
\textbf{L}egendre
\textbf{M}emory model, or \textbf{FiLM} for short, for long-term time series forecasting.

In short, we summarize the key contributions of this work as follows:
\begin{enumerate}
    \item We propose a {\it Frequency improved Legendre Memory model (FiLM)} architecture with a mixture of experts for robust multiscale time series feature extraction.
    \item We redesign the {\it Legendre Projection Unit (LPU)} and make it a general tool for data representation that any time series forecasting model can exploit to solve the historical information preserving problem.
    \item We propose {\it Frequency Enhanced Layers (FEL)} that reduce dimensionality by a combination of Fourier analysis and low-rank matrix approximation to minimize the impact of noisy signals from time series and ease the overfitting problem. The effectiveness of this method is verified both theoretically and empirically.
    \item We conduct extensive experiments on six benchmark datasets across multiple domains (energy, traffic, economics, weather, and disease). Our empirical studies show that the proposed model improves the performance of state-of-the-art methods by \textbf{19.2\%} and \textbf{26.1\%} in multivariate and univariate forecasting, respectively. In addition, our empirical studies also reveal a dramatic improvement in computational efficiency through dimensionality reduction. 
\end{enumerate}

\section{Time Series Representation in Legendre-Fourier Domain} \label{sec:theoretical}


\subsection{Legendre Projection}
 Given an input sequence, the function approximation problem aims to approximate the cumulative history at every time $t$. Using Legendre Polynomials projection, we can project a prolonged sequence of data onto a subspace of bounded dimension, leading to compression, or feature representation, for evolving historical data. Formally, given a smooth function $f$ observed online, we aim to maintain a fixed size compressed representation of its history $f(x)_{[t-\theta,t]}$, where $\theta$ specifies the window size. At every time point $t$, the approximation function $g^{(t)}(x)$ is defined with respect to the measure $\mu^{(t)} =\frac{1}{\theta}\mathbbm{I}_{[t-\theta,t]}(x)$. In this paper, we use Legendre Polynomials of degree at most $N-1$ to build the function $g^{(t)}(x)$, i.e.
  \begin{align}
  g^{(t)}(x)=\sum_{n=1}^N c_{n}(t) P_{n}\left(\frac{2(x-t)}{\theta}+1\right), \label{eq:1}
 \end{align}
 where $P_n(\cdot)$ is the $n$-th order Legendre Polynomials. Coefficients $c_{n}(t)$ are captured by the following dynamic equation:
\begin{align}
 \frac{d}{d t} c(t) &=-\frac{1}{\theta} A c(t)+\frac{1}{\theta} B f(t).
\end{align}
where the definition of $A$ and $B$ can be found in~\cite{LMU}. Using Legendre Polynomials as a basis allows us to accurately approximate smooth functions, as indicated by the following theorem.

\begin{theorem} [\it Similar to Proposition 6 in \cite{Hippo}]
 If $f(x)$ is $L$-Lipschitz, then $\|f_{[t-\theta,t]}(x)-g^{(t)}(x)\|_{\mu^{(t)}}\le \mathcal{O}(\theta L/\sqrt{N})$. Moreover, if $f(x)$ has $k$-th order bounded derivatives, we have $\|f_{[t-\theta,t]}(x)-g^{(t)}(x)\|_{\mu^{(t)}}\le \mathcal{O}(\theta^kN^{-k+1/2})$.
\end{theorem}
According to Theorem 1, without any surprise, the larger the number of Legendre Polynomials basis, the more accurate the approximation will be, which unfortunately may lead to the overfitting of noisy signals in the history. As shown in Section \ref{sec_exp}, directly feeding deep learning modules, such as MLP, RNN, and vanilla Attention without modification, with the above features will not yield state-of-the-art performance, primarily due to the noisy signals in history. That is why we introduce, in the following subsection, a Frequency Enhanced Layer with Fourier transforms for feature selection.  


Before we close this subsection, we note that unlike~\cite{S4}, a fixed window size is used for function approximation and feature extraction. This is largely because a longer history of data may lead to a larger accumulation of noises from history. To make it precise, we consider an auto-regressive setting with random noise. Let $\{x_t\}\in \mathbbm{R}^d$ be the time sequence with $x_{t+1} = Ax_t + b + \epsilon_t$ for $t=1,2,...$, where $A\in\mathbbm{R}^{d\times d}$, $b \in \mathbbm{R}^d$, and $\epsilon_t\in\mathbbm{R}^{d}$ is random noise sampled from $N(0, {\sigma^2 I})$. As indicated by the following theorem, given $x_t$, the noise will accumulate in $x_{t-\theta}$ over time at the rate of $\sqrt{\theta}$, where $\theta$ is the window size.

\begin{theorem}
Let $A$ be an unitary matrix and $\epsilon_t$ be $\sigma^2$-subgaussian random noise.  We have $x_{t} = A^{\theta}x_{t-\theta} + \sum_{i=1}^{\theta-1}A^{i}b + \mathcal{O}(\sigma \sqrt{\theta})$.
\end{theorem}



\subsection{Fourier Transform}
Because white noise has a completely flat power spectrum, it is commonly believed that the time series data enjoys a particular spectral bias, which is not generally randomly distributed in the whole spectrum.
Due to the stochastic transition environment, the real output trajectories of the forecasting task contain large volatilises and people usually only predict the mean path of them. The relative more smooth solutions thus are preferred. According to Eq.~\eqref{eq:1},  the approximation function $g^{(t)}(x)$ can be stabilized via smoothing the coefficients $c_n(t)$ in both $t$ and $n$. This observation helps us design an efficient data-driven way to tune the coefficients $c_n(t)$. As smoothing in $n$ can be simply implemented via multiplying learnable scalars to each channel, we mainly discuss smoothing $c_n(t)$ in $t$ via Fourier transformation. The spectral bias implies the the spectrum of $c_n(t)$ mainly locates in low frequency regime and has weak signal strength in high frequency regime. To simplify our analysis, let us assume the Fourier coefficients of $c_{n}(t)$ as $a_{n}(t)$. Per spectral bias, we assume that there exists a $s, a_{\min}>0$, such that for all $n$ we have $t> s$, $|a_{n}(t)|\le a_{\min}$. An idea to sample coefficients is to keep the the first $k$ dimensions and randomly sample the remaining dimensions instead of the fully random sampling strategy. We characterize the approximation quality via the following theorem:

\begin{theorem}
\label{thm_Fourier} 
Let $A\in\mathbbm{R}^{d\times n}$ be the Fourier coefficients matrix of an input matrix $X\in\mathbbm{R}^{d\times n}$, and $\mu(A)$, the coherence measure of matrix $A$, is $\Omega(k/n)$. We assume there exist $s$ and  a positive $ a_{\min}$ such that the elements in last $d-s$ columns of $A$ is smaller than $a_{\min}$. If we keep first $s$ columns selected and randomly choose $\mathcal{O}(k^2/\epsilon^2 - s)$ columns from the remaining parts, with high probability
\[
    \|A - P(A)\|_F \leq \mathcal{O}\left[(1 + \epsilon)a_{\min}\cdot\sqrt{(n-s)d}\right],
\]
where  $P(A)$ denotes the matrix projecting $A$ onto the column selected column space.
\end{theorem}
Theorem~\ref{thm_Fourier} implies that when $a_{\min}$ is small enough, the selected space can be considered as almost the same as the original one.
\section{Model Structure}\label{sec_model}
\begin{figure*}[t]
\centering
\includegraphics[width=0.9\linewidth]{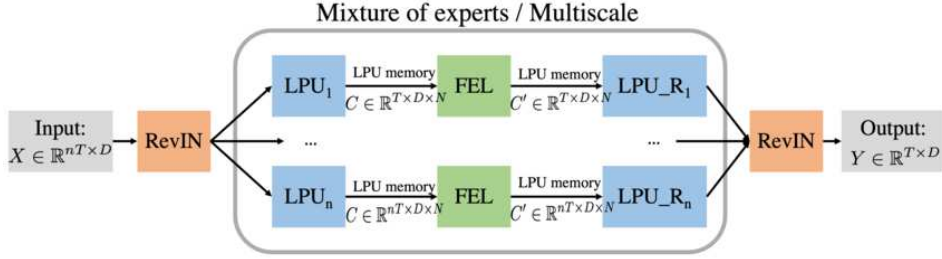}
\caption{Overall structure of FilM (Frequency improved Legendre Memory Model). LPU: Legendre Projection Unit. LPU\_R: reverse recovery of Legendre Projection. FEL: Frequency Enhanced Layer. RevIn: data normalization block. The input data is first normalized and then project to Legendre Polynomials space (LPU memory $C$). The LPU memory $C$ is processed with FEL and generates the memory $C'$ of output. Finally, $C'$ is reconstructed and denormalized to output series with $\mathrm{LPU}_R$. A multiscale structure is employed to process the input with length $\{T,\ 2T,\ ...\ nT\}$.} 
\label{fig:film_overall_strcuture}
\vskip -0.10in
\end{figure*}

\subsection{FiLM: Frequency improved Legendre Memory Model}\label{sec:model_structure}
The overall structure of FiLM is shown in Figure \ref{fig:film_overall_strcuture}
The FiLM maps a sequence $X \mapsto Y$, where $X, Y \in \R^{T \times D}$, by mainly utilizing two sub-layers: Legendre Projection Unit (LPU) layer and Fourier Enhanced Layer (FEL). In addition, to capture history information at different scales, a mixture of experts at different scales is implemented in the LPU layer. An optional add-on data normalization layer RevIN \cite{reversible} is introduced to further enhance the model's robustness. It is worth mentioning that FiLM is a simple model with only one layer of LPU and one layer of FEL.

\begin{figure}[t]
\begin{minipage}{\linewidth}
    \centering
    \scalebox{0.7}{
    \includegraphics[width=\linewidth]{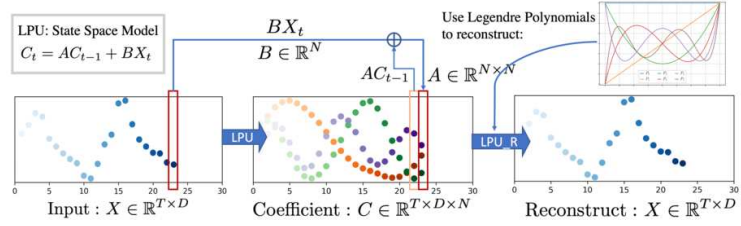}}
    \end{minipage}
    \caption{Structure of Legendre Projection Layer (LPU). LPU contains two states: Projection \& Reconstruction. $C(t)$ is the compressed memory for historical input up to time $t$. $x(t)$ is the original input signal at time $t$. A, B are two pre-fixed projection matrices. $C(t)$ is reconstructed to original input by multiplying a discrete Legendre Polynomials matrix.}
    \label{fig:LPL}
\vskip -0.1in
\end{figure}

\paragraph{LPU: Legendre Projection Unit} LPU is a state space model: $C_t = AC_{t-1} + Bx_t$, where $x_t \in \R$ is the input signal, $C_t \in \R^N$ is the memory unit, and $N$ is the number of Legendre Polynomials. LPU contains two untrainable prefixed matrices $A$ and $B$ defined as follows:
\begin{align}
A_{n k}=(2 n+1) \begin{cases}(-1)^{n-k} & \text { if } k \leq n \\ 1 & \text { if } k \geq n\end{cases}, 
B_{n}=(2 n+1)(-1)^{n}. 
\label{equ:LPU_A_B}
\end{align}
LPU contains two stages, i.e., Projection and Reconstruction. The former stage projects the original signal to the memory unit: $C=\mathrm{LPU}(X)$. The later stage reconstructs the signal from the memory unit: $X_{re}=\mathrm{LPU\_R}(C)$. The whole process in which the input signal is projected/reconstructed to/from memory $C$ 
is shown in Figure \ref{fig:LPL}.

\paragraph{FEL: Frequency Enhanced Layer} 
\begin{figure}[t]
    \begin{minipage}{\linewidth}
     \centering
     \scalebox{0.70}{
    \includegraphics[width=\linewidth]{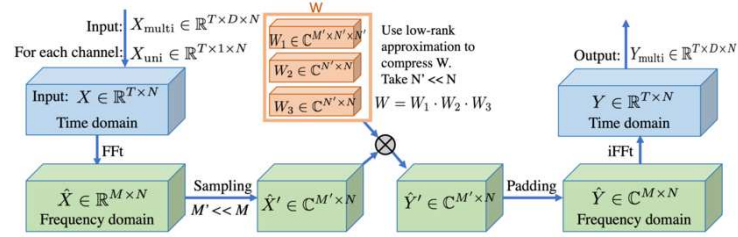}}
    \end{minipage}
    \caption{Structure of Frequency Enhanced Layer (FEL): Original version (use weights $W$) and Low-rank Approximation version (use weights $W=W_1 \cdot W_2 \cdot W_3$), where N is Legendre Polynomials number, M is Fourier mode number, and T is sequence length.}
    \label{fig:FEL}
\vskip -0.1in
\end{figure}

\subparagraph{Low-rank Approximation} The FEL is with a single learnable weight matrix ($W \in \mathbb{R}^{M' \times N \times N}$) which is all we need to learn from the data. However, this weight could be large. We can decompose $W$ into three matrices $W_1 \in \mathbb{R}^{M' \times N' \times N'}$, $W_2 \in \mathbb{R}^{N' \times N}$ and $W_3 \in \mathbb{R}^{N' \times N}$ to perform a low-rank approximation ($N' << N$). Take Legendre Polynomials number $N=256$ as default, our model's learnable weight can be significantly reduced to \textbf{0.4\%} with $N'=4$ with minor accuracy deterioration as shown in Section~\ref{sec_exp}. The calculation mechanism is described in Figure \ref{fig:FEL}.

\subparagraph{Mode Selection}
A subset of the frequency modes is selected after Fourier transforms to reduce the noise and boost the training speed. Our default selection policy is choosing the lowest M mode. Various selection policies are studied in the experiment section. Results show that adding some random high-frequency modes can give extra improvement in some datasets, as supported by our theoretical studies in Theorem~\ref{thm_Fourier}.

Implementation source code for LPU and FEL are given in Appendix \ref{app:Alg}.

\subsection{Mixture of Multiscale Experts Mechanism}\vspace{-1mm}
Multiscale phenomena is a unique critical data bias for time series forecasting. Since we treat history sequence points with uniform importance, our model might lack such prior. Our model implemented a simple mixture of experts strategy that utilizes the input sequence with various time horizons $\{T,\ 2T,\ ...\ nT\}$ to forecast the predicted horizon $T$ and merge each expert prediction with linear layer as show in Figure \ref{fig:film_overall_strcuture}. This mechanism improves the model's performance consistently across all datasets, as shown in Table \ref{tab:ablation_normalize}.


\subsection{Data Normalization}\vspace{-1mm}
As \cite{Autoformer,FedFormer} point out, time series seasonal-trend decomposition is a crucial data normalization design for long-term time series forecasting. We find that our LMU projection can inherently play a normalization role for most datasets, but lacking an explicate normalization design still hurt the robustness of performance in some cases. A simple reversible instance normalization (RevIN) \cite{reversible} is adapted to act as an add-on explicate data normalization block. 
The mean and standard deviation are computed for every instance 
$x_{k}^{(i)} \in \mathbb{R}^{T}$ of the input data as
\scalebox{0.8}{$
\mathbb{E}_{t}\left[x_{k t}^{(i)}\right]=\frac{1}{T} \sum_{j=1}^{T} x_{k j}^{(i)} \quad \text { and } \quad \operatorname{Var}\left[x_{k t}^{(i)}\right]=\frac{1}{T} \sum_{j=1}^{T}\left(x_{k j}^{(i)}-\mathbb{E}_{t}\left[x_{k t}^{(i)}\right]\right)^{2}.
$}
Using these statistics, we normalize the input data $x^{(i)}$  as
\scalebox{0.8}{$
\hat{x}_{k t}^{(i)}=\gamma_{k}\left(\frac{x_{k t}^{(i)}-\mathbb{E}_{t}\left[x_{k t}^{(i)}\right]}{\sqrt{\operatorname{Var}\left[x_{k t}^{(i)}\right]+\epsilon}}\right)+\beta_{k},
$}
where $\gamma, \beta \in \mathbb{R}^{K}$ are learnable affine parameter vectors. 
Then the normalized input data is sent into the model for forecasting. In the end, we denormalize the model output by applying the reciprocal of the normalization performed at the beginning. 

RevIN slows down the training process by 2-5 times, and we do not observe consistent improvement on all datasets by applying RevIn. Thus, it can be considered an optional stabilizer in model training. Its detailed performance is shown in the ablation study in Table \ref{tab:ablation_normalize}.


\section{Experiments}\label{sec_exp}\vspace{-2mm}

\begin{table*}[h]
\centering
\caption{multivariate long-term series forecasting results on six datasets with various input length and prediction length $O \in \{96,192,336,720\}$ (For ILI dataset, we set prediction length $O \in \{24,36,48,60\}$). A lower MSE indicates better performance. All experiments are repeated 5 times.}\vspace{-1mm}
\scalebox{0.60}{
\begin{tabular}{c|c|cccccccccccccccccc}
\toprule
\multicolumn{2}{c|}{Methods}&\multicolumn{2}{c|}{FiLM}&\multicolumn{2}{c|}{FEDformer}&\multicolumn{2}{c|}{Autoformer}&\multicolumn{2}{c|}{S4}&\multicolumn{2}{c|}{Informer}&\multicolumn{2}{c|}{LogTrans}&\multicolumn{2}{c}{Reformer}\\
\midrule
\multicolumn{2}{c|}{Metric} & MSE  & MAE & MSE & MAE& MSE  & MAE& MSE  & MAE& MSE  & MAE& MSE  & MAE& MSE  & MAE\\
\midrule
\multirow{4}{*}{\rotatebox{90}{$ETTm2$}} &96 & \textbf{0.165} & \textbf{0.256} &0.203 &0.287 &0.255  &0.339 &0.705 &0.690 &0.365  &0.453  &0.768  &0.642  &0.658  &0.619    \\
                        & 192 & \textbf{0.222} & \textbf{0.296} &0.269  &0.328  &0.281 &0.340 &0.924 &0.692 &0.533  &0.563  &0.989  &0.757  &1.078  &0.827    \\
                        & 336 & \textbf{0.277} & \textbf{0.333} & 0.325 &0.366 &0.339  &0.372 &1.364 &0.877 &1.363&0.887  &1.334  &0.872  &1.549  &0.972     \\
                        & 720 & \textbf{0.371} & \textbf{0.389} &0.421 &0.415 &0.422  &0.419 &0.877 &1.074 &3.379  &1.338 & 3.048 &1.328  &2.631  &1.242      \\
\midrule
\multirow{4}{*}{\rotatebox{90}{$Electricity$}} &96  &\textbf{0.154}  &\textbf{0.267}  &0.183 &0.297 &0.201  &0.317 &0.304 &0.405 &0.274  &0.368  &0.258  &0.357  &0.312  &0.402    \\
                        & 192 &\textbf{0.164} & \textbf{0.258} & 0.195 &0.308 &0.222  &0.334 &0.313 &0.413 &0.296 &0.386  &0.266 &0.368  &0.348  &0.433    \\
                        & 336 & \textbf{0.188} & \textbf{0.283} &0.212 &0.313 &0.231 &0.338 &0.290 &0.381 &0.300  &0.394  &0.280 &0.380  &0.350  & 0.433    \\
                        & 720 & 0.236 & \textbf{0.332} &\textbf{0.231} &0.343 &0.254  &0.361 &0.262 &0.344 &0.373  &0.439 &0.283  &0.376  &0.340  &0.420     \\
\midrule
\multirow{4}{*}{\rotatebox{90}{$Exchange$}} &96  & \textbf{0.086} & \textbf{0.204} &0.139 &0.276 &0.197  &0.323 &1.292 &0.849 &0.847  &0.752  &0.968  &0.812  &1.065  &0.829    \\
                        & 192 & \textbf{0.188} & \textbf{0.292} &0.256 &0.369 &0.300  &0.369 &1.631 &0.968 &1.204   &0.895 &1.040  &0.851  &1.188  & 0.906   \\
                        & 336 & \textbf{0.356} & \textbf{0.433} &0.426 &0.464 &0.509  &0.524 &2.225 &1.145 &1.672  &1.036  &1.659  &1.081  &1.357  &0.976     \\
                        & 720 & \textbf{0.727} & \textbf{0.669} &1.090 &0.800 &1.447  &0.941 &2.521 &1.245 &2.478  &1.310  &1.941  &1.127  &1.510  &1.016     \\
\midrule
\multirow{4}{*}{\rotatebox{90}{$Traffic$}} &96  &\textbf{0.416}& \textbf{0.294} &0.562 &0.349 &0.613  &0.388 &0.824 &0.514 &0.719  &0.391  &0.684  &0.384  &0.732  &0.423    \\
                        & 192 & \textbf{0.408} & \textbf{0.288} &0.562 &0.346 &0.616&0.382 &1.106 &0.672 &0.696 &0.379  &0.685  &0.390  &0.733  &0.420    \\
                        & 336 & \textbf{0.425} & \textbf{0.298} &0.570 &0.323 &0.622  &0.337 &1.084 &0.627 &0.777  &0.420  &0.733  &0.408  &0.742  &0.420     \\
                        & 720 & \textbf{0.520} &\textbf{0.353} &0.596 &0.368 &0.660  &0.408 &1.536 &0.845 &0.864  &0.472  &0.717  &0.396  &0.755  &0.423     \\
\midrule
\multirow{4}{*}{\rotatebox{90}{$Weather$}} & 96 & \textbf{0.199} & \textbf{0.262} &0.217  &0.296  &0.266  &0.336 &0.406 &0.444 &0.300  &0.384  &0.458  &0.490  &0.689  &0.596    \\
                        & 192 & \textbf{0.228} & \textbf{0.288} &0.276  &0.336  &0.307  &0.367 &0.525 &0.527 &0.598  &0.544  &0.658  &0.589  &0.752  &0.638    \\
                        & 336 &\textbf{0.267} &\textbf{0.323} & 0.339  &0.380  &0.359  &0.395 &0.531 &0.539 &0.578  &0.523  &0.797  &0.652  &0.639  &0.596    \\
                        & 720 &\textbf{0.319} & \textbf{0.361} &0.403  &0.428 &0.578 &0.578 &0.419  &0.428  &1.059  &0.741  &0.869  &0.675  &1.130  &0.792    \\
\midrule
\multirow{4}{*}{\rotatebox{90}{$ILI$}} & 24 & \textbf{1.970} & \textbf{0.875} &2.203  &0.963  &3.483 &1.287 &4.631 &1.484 &5.764  &1.677  &4.480  &1.444  &4.400 &1.382    \\
                        & 36 & \textbf{1.982} & \textbf{0.859} &2.272  &0.976  &3.103  &1.148 &4.123 &1.348 &4.755  &1.467  &4.799  &1.467  &4.783  &1.448    \\
                        & 48 & \textbf{1.868} & \textbf{0.896} &2.209  &0.981  &2.669  &1.085 &4.066 &1.36 &4.763  &1.469  &4.800  &1.468  &4.832  &1.465    \\
                        & 60 & \textbf{2.057} & \textbf{0.929} &2.545  &1.061  &2.770  &1.125 &4.278 &1.41 &5.264  &1.564  &5.278  &1.560  &4.882  &1.483    \\
\bottomrule
\end{tabular}
\label{tab:multi-benchmarks}
}
\end{table*}

To evaluate the proposed FiLM, we conduct extensive experiments on six popular real-world benchmark datasets for long-term forecasting, including traffic, energy, economics, weather, and disease. 
Since classic models such as ARIMA and simple RNN/TCN models perform inferior as shown in \cite{haoyietal-informer-2021} and \cite{Autoformer}, we mainly include five state-of-the-art (SOTA) Transformer-based models, i.e., FEDformer, Autoformer~\cite{Autoformer}, Informer~\cite{haoyietal-informer-2021}, LogTrans~\cite{Log-transformer-shiyang-2019}, Reformer~\cite{DBLP:conf/iclr/KitaevKL20-reformer}, and one recent state-space model with recursive memory S4~\cite{S4}, for comparison. FEDformer is selected as the main baseline as it achieves SOTA results in most settings. More details about baseline models, datasets, and implementations are described in Appendix. 

\subsection{Main Result}
For better comparison, we follow the experiment settings of Informer \cite{haoyietal-informer-2021} where the input length is tuned for best forecasting performance, and the prediction lengths for both training and evaluation are fixed to be 96, 192, 336, and 720, respectively.

\paragraph{Multivariate Results}
In multivariate forecasting tasks,
FiLM achieves the best performance on all six benchmark datasets at all horizons, as shown in Table \ref{tab:multi-benchmarks}. Compared with SOTA work (FEDformer), our proposed FiLM yields an overall \textbf{20.3\%} relative MSE reduction. It is worth noting that the improvement is even more significant on some of the datasets, such as Exchange (over 30\%). The Exchange dataset does not exhibit apparent periodicity, but FiLM still achieves superior performance.  
The improvements made by FiLM are consistent with varying horizons, demonstrating
its strength in long-term forecasting. More results
on the ETT full benchmark are provided in Appendix \ref{app:exp:ett_benchmark}.

\paragraph{Univariate Results}
The benchmark results for univariate time series
forecasting are summarized in Appendix \ref{app:exp:uni_bench}, Table \ref{tab:uni-benchmarks-large}. Compared with
SOTA work (FEDformer), FiLM yields an overall \textbf{22.6\%} relative
MSE reduction. And on some datasets, such as Weather and
Electricity, the improvement can reach more than 40\%. It again
proves the effectiveness of FiLM in long-term forecasting.

\begin{figure*}[t]
\centering
\includegraphics[width=0.8\linewidth]{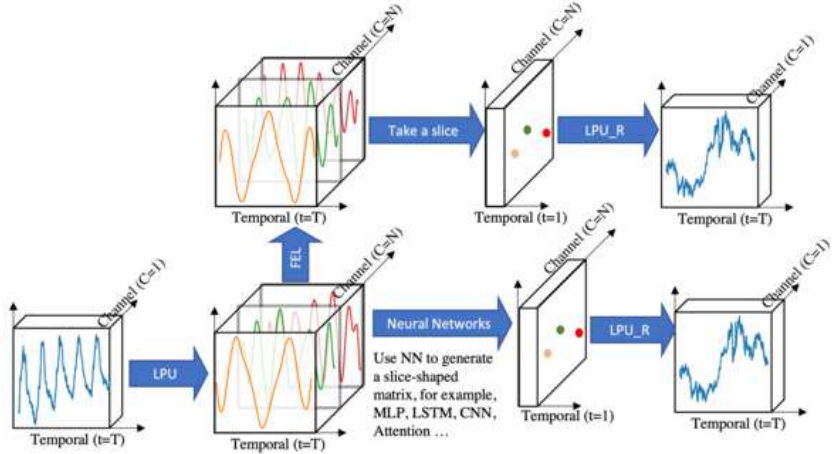}
\caption{LPU boosting effect. LPU can serve as a plug-in block in various backbones, e.g., FEL, MLP, LSTM, CNN, and Attention. Replacing LPU with a comparable-sized linear layer will always lead to degraded performance.}
\label{fig:boost}
\vskip -0.2in
\end{figure*}

\paragraph{LPU Boosting Results}
\vskip -0.1in

\begin{wraptable}[12]{r}{0.7\textwidth}
\centering
\caption{Boosting effect of LPU layer for common deep learning backbones: MLP, LSTM, CNN and Attention.`+` indicates degraded performance.}\vspace{-1mm}
\scalebox{0.60}{
\begin{tabular}{c|c|cccccccccccccccccc}
\toprule
\multicolumn{2}{c|}{Methods}&\multicolumn{2}{c|}{FEL}&\multicolumn{2}{c|}{MLP}&\multicolumn{2}{c|}{LSTM}&\multicolumn{2}{c|}{lagged-LSTM}&\multicolumn{2}{c|}{CNN}&\multicolumn{2}{c}{Attention}\\
\midrule
\multicolumn{2}{c|}{Compare} & LPU  & Linear& LPU  & Linear & LPU  & Linear & LPU  & Linear & LPU  & Linear & LPU  & Linear\\
\midrule

\multirow{4}{*}{\rotatebox{90}{ETTm1}} 

&96  &\textbf{0.030}  & +38\%   & 0.034  & +8.0\%  & 0.049 & +73\% & 0.093 & -21\% & 0.116   & -50\%   & 0.243  & -81\% \\
& 192  &\textbf{0.047} & +9.5\% & 0.049  & +30\%  & 0.174 & +32\% & 0.331 & -48\% & 0.101   & +20\%    & 0.387  & -86\% \\
& 336 & \textbf{0.063} & +5.8\% & 0.061  & +64\%  & 0.119 & +84\% & 0.214 & -19\% & 0.122   & +25\%    & 1.652  & +12\% \\
& 720 &\textbf{0.081} & +1.4\%  & 0.082  & +62\%  & 0.184 & +32\% & 0.303 & -6.5\%  & 0.108   & +13\%    & 4.782  & -61\% \\

\midrule
\multirow{4}{*}{\rotatebox{90}{Electricity}} 

&96  &\textbf{0.213}  & +136\%  & 0.431  & +121\%   & 0.291 & +55.6\% & 0.739 & -33\% & 0.310  & +43\%  & 0.805 & +23\% \\
& 192 &\textbf{0.268} & +32\%   & 0.291 & +239\%   & 0.353  & +17\% & 0.535 & +15\% & 0.380  & +12\%  & 0.938 & +14\% \\
& 336 & \textbf{0.307}& +0.1\%  & 0.296 & +235\%   & 0.436  & -6.7\% & 0.517 & +23\% & 0.359  & +29\% & 2.043 & -54\% & \\
& 720 & \textbf{0.321}& +37\%   & 0.339 & +196\%   & 0.636  & -11\% & 0.492 & +28\% & 0.424  & +18\%  & 9.115 & +298\% \\ 

\bottomrule
\end{tabular}
\label{tab:boosting}
}
\end{wraptable}
A set of experiments are conducted to measure the boosting effect of LPU when combined with various common deep learning modules (MLP, LSTM, CNN, and Attention), as shown in Figure \ref{fig:boost}. In the experiment, we compare the LPU and the comparable-sized linear layer. It is worth mentioning that LPU does not contain any learnable parameter. 
The results are shown in Table \ref{tab:boosting}. For all the modules, LPU significantly improves their average performance for long-term forecasting: MLP: \textbf{119.4\%}, LSTM: \textbf{97.0\%}, CNN: \textbf{13.8\%}, Attention: \textbf{8.2\%}. Vanilla Attention has a relative poor performance when combining the LPU, which worth further digging.

\paragraph{Low-rank Approximation for FEL}

\begin{wraptable}[11]{r}{0.6\textwidth}
\vskip -0.15in
\centering
\caption{Low-rank Approximation (LRA) study for Frequency Enhanced Layer (FEL): Comp. K=0 means default version without LRA, 1 means the largest compression using K=1.}\vspace{-1.5mm}
\scalebox{0.65}{
\begin{tabular}{c|c|cccccccccc}
\toprule

\multicolumn{2}{c|}{Comp. K}&\multicolumn{2}{c|}{0}&\multicolumn{2}{c|}{16}&\multicolumn{2}{c|}{4}&\multicolumn{2}{c}{1}\\

\midrule
\multicolumn{2}{c|}{Metric} & MSE  & MAE & MSE & MAE& MSE  & MAE& MSE  & MAE \\
\midrule


\multirow{4}{*}{\rotatebox{90}{ETTh1}} 

&96  &\textbf{0.371} &\textbf{0.394}    &0.371  &0.396  &0.371  &0.398   & 0.400   & 0.421    \\
& 192  &0.414 &\textbf{0.423}  &0.411  &0.423  &0.414  &0.426   & 0.435   & 0.444    \\
& 336 & \textbf{0.442} & 0.445 &0.443  &0.446  &0.443  & 0.444  & 0.492   &  0.478     \\
& 720 &\textbf{0.454}& \textbf{0.451}   &0.464  &0.474  &0.468  &0.478   & 0.501   & 0.499      \\

\midrule

\multirow{4}{*}{\rotatebox{90}{Weather}} 
&96  &\textbf{0.199} &\textbf{0.262}   &0.199  &0.263  &0.197  &0.262   &0.198    & 0.263     \\
& 192  &0.228 &0.288  &0.225  &0.285  &0.226  &0.285   &0.225    & 0.286      \\
& 336 & 0.267 & 0.323 &0.266  &0.321  &0.263  &0.314   & 0.264   &  0.316      \\
& 720 &0.319& 0.361  &0.314 &0.355  &0.315  & 0.354   &0.318    & 0.357   \\
\midrule


\multicolumn{2}{c|}{Parameter size}&\multicolumn{2}{c|}{100\%}&\multicolumn{2}{c|}{1.95\%}&\multicolumn{2}{c|}{0.41\%}&\multicolumn{2}{c}{0.10\%}\\
\bottomrule
\end{tabular}
\label{tab:FEL_LRA}
}
\end{wraptable}
Low-rank approximation of learnable matrix in Frequency Enhanced Layer can significantly reduce our parameter size to \textbf{0.1\%$\sim$0.4\%} with minor accuracy deterioration. The experiment details are shown in Table \ref{tab:FEL_LRA}. Compared to Transformer-based baselines, FiLM enjoys \textbf{80\%} learnable parameter reduction and \textbf{50\%} memory usage reductions, as shown in Appendix \ref{app:sec:learnable_para_size}, \ref{app:time}.  

\paragraph{Mode selection policy for FEL}

\begin{wraptable}[11]{r}{0.5\textwidth}
\vskip -0.1in

\centering
\caption{Mode selection policy study for frequency enhanced layer. Lowest: select the lowest $m$ frequency mode; Random: select $m$ random frequency mode; Low random: select the $0.8*m$ lowest frequency mode and $0.2*m$ random high frequency mode.}\vspace{-1.5mm}
\scalebox{0.65}{
\begin{tabular}{c|c|cccccccc}
\toprule
\multicolumn{2}{c|}{Policy}&\multicolumn{2}{c|}{Lowest}&\multicolumn{2}{c|}{Random}&\multicolumn{2}{c}{Low random}\\
\midrule
\multicolumn{2}{c|}{Metric} & MSE  & MAE & MSE & MAE& MSE  & MAE \\
\midrule

\multirow{4}{*}{\rotatebox{90}{Exchange}} 

&96  &\textbf{0.086} &\textbf{0.204}    &0.086  &0.208  &0.087  &0.210        \\
& 192  &0.188 &\textbf{0.292}  &\textbf{0.187}  &0.318  &0.207  &0.340      \\
& 336 & 0.356 & \textbf{0.433} &0.358  &0.437  &\textbf{0.353}  &0.461   \\
& 720 &\textbf{0.727}& \textbf{0.669}   &0.788  &0.680  &0.748  &0.674   \\

\midrule

\multirow{4}{*}{\rotatebox{90}{Weather}} 

&96  &0.199 &0.262 &0.197  &0.256  &\textbf{0.196}  &\textbf{0.254}   \\
& 192  &\textbf{0.228} &\textbf{0.288}  &0.234  &0.300  &0.234  &0.301   \\
& 336 & 0.267 & 0.323 &0.266  &0.319  &\textbf{0.263}  &\textbf{0.316}   \\
& 720 &0.319& 0.361 &0.317  &0.356  &\textbf{0.316}  &\textbf{0.354}  \\

\bottomrule

\end{tabular}
\label{tab:mode}
}
\end{wraptable}

Frequency mode selection policy is studied in Table \ref{tab:mode}. The \textit{Lowest} mode selection method shows the most robust performance. The results in \textit{Low random} column shows that randomly adding some high-frequency signal gives extra improvement in some datasets, as our theoretical studies in Theorem \ref{thm_Fourier} support.


\subsection{Ablation Study}\label{subsec:ablation}

This subsection presents the ablation study of the two major blocks (FEL \& LPU) employed, the multiscale mechanism, and data normalization (RevIN). 

\paragraph{Ablation of LPU}
\begin{table*}[h]

\vskip -0.05in
\centering
\caption{Ablation studies of LPU layer. The original LPU block (whose projection and reconstruction matrix are fixed) is replaced with 6 variants (\textbf{Fixed} means the matrix are not trainable. \textbf{Trainable} means the matrix is initialized with original parameters and trainable. \textbf{Random Init} means the matrix is initialized randomly and trainable). The experiments are performed on ETTm1 and Electricity with different input lengths. The metric of variants is presented in relative value (`+` indicates degraded performance, `-` indicates improved performance).}
\vspace{-1mm}
\resizebox{\columnwidth}{!}{
\scalebox{0.60}{

\begin{tabular}{c|c|cccccccccccccccccc}
\toprule

\multicolumn{2}{c|}{Index}&\multicolumn{2}{c|}{Original}&\multicolumn{2}{c|}{Variant 1}&\multicolumn{2}{c|}{Variant 2}&\multicolumn{2}{c|}{Variant 3}&\multicolumn{2}{c|}{Variant 4}&\multicolumn{1}{c|}{Variant 5}&\multicolumn{2}{c}{Variant 6}\\
\midrule

\multicolumn{2}{c|}{Projection}&\multicolumn{4}{c|}{Fixed}&\multicolumn{6}{c|}{Trainable}&\multicolumn{1}{c|}{Random Init}&\multicolumn{2}{c}{Linear}\\
\midrule
\multicolumn{2}{c|}{Reconstruction}&\multicolumn{2}{c|}{Fixed}&\multicolumn{2}{c|}{Trainable}&\multicolumn{2}{c|}{Fixed}&\multicolumn{2}{c|}{Trainable}&\multicolumn{2}{c|}{Random Init}&\multicolumn{1}{c|}{--}&\multicolumn{2}{c}{Linear}\\

\midrule
\multicolumn{2}{c|}{Metric} & MSE  & MAE & MSE & MAE& MSE  & MAE& MSE  & MAE& MSE  & MAE& -- & MSE  & MAE \\
\midrule
\multirow{4}{*}{\rotatebox{90}{ETTm1}} 

&96  &\textbf{0.030} &\textbf{0.128}    &+6.0\%  & +3.1\% & +4.6\% & +3.1\%  & +4.7\%   & +2.7\%   & +0.7\% & +0.6\%  & NaN & +38\% & +22\%  \\
& 192  &\textbf{0.047} &\textbf{0.160}  &-2.6\%  & -1.3\%    & +2.5\% & +1.8\%  & -0.4\%   & +0.4\%   & -3.0\% & -1.0\%  & NaN  & +9.5\%  & +8.7\% \\
& 336 & \textbf{0.063} & \textbf{0.185} &+4.2\% & +1.8\%    & -2.4\% & -0.4\%  & -6.3\%   & -2.5\%    & +2.2\% & +1.4\%   & NaN & +5.8\% & +5.0\% \\
& 720 &\textbf{0.081}& \textbf{0.215}   &-0.4\%     & -0.6\%     &+24\%    & +12\%  & +11\%    & +5.9\%   & NaN & NaN       & NaN      & +1.4\%  & +2.2\% \\

\midrule

\multirow{4}{*}{\rotatebox{90}{Electricity}}
&96  &\textbf{0.213}  &\textbf{0.328}  &+15\%  & +6.8\% & +11\% & +5.2\%  & +11\%    & +5.1\% & +11\% & +4.8\%  &NaN & +136\% & +58\%  \\
& 192 &\textbf{0.268} & \textbf{0.373} &-4.6\% & -3.8\% & +7.8\%  & +3.9\%   & +6.8\%    & +3.5\% & -5.3\% & -3.9\%  &NaN  & +32\%  & +16\% \\
& 336 & \textbf{0.307}& \textbf{0.417}&-4.2\%  & -5.0\% & -3.9\%  & -6.0\%   & -7.2\%    & -8.0\%   & -8.5\% & -9.0\%   &NaN & +0.1\% & -5.0\% \\
& 720 & \textbf{0.321}&\textbf{0.423} &+2.9\%  & +2.8\% & +10\%  &  +6.8\%  & +3.0\%   & +1.7\%  & +207\% & +85\%      &NaN  & 37\%  & 22\% \\

\bottomrule
\end{tabular}
\label{tab:ablation_HiPPO}
}}
\vskip -0.05in
\end{table*}

To prove the effectiveness of LPU layer, in Table \ref{tab:ablation_HiPPO}, we compare the original LPU layer with six variants. The LPU layer consists of two sets of matrices (Projection matrices \& Reconstruction matrices). Each has three variants: Fixed, Trainable, and Random Init. In Variant 6, we use comparable-sized linear layers to replace the LPU layer. We observe that Variant 6 leads to a 32.5\% degradation on average, which confirms the effectiveness of Legendre projection. The projection matrix in LPU is recursively called $N$ times ($N$ is the input length). So if the projection matrix is randomly initialized (Variant 5), the output will face the problem of exponential explosion. If the projection matrix is trainable (Variants 2, 3, 4), the model suffers from the exponential explosion as well and thus needs to be trained with a small learning rate, which leads to a plodding training speed and requires more epochs for convergence. Thus, the trainable projection version is not recommended, considering the trade-off between speed and performance. The variant with trainable reconstruction matrix (Variant 1) has comparable performance and less difficulty in convergence.

\paragraph{Ablation of FEL}


\begin{table*}[ht]
\vskip -0.05in
\centering
\caption{Ablation studies of FEL layer. The FEL layer is replaced with 4 different variants: MLP, LSTM, CNN, and Transformer. S4 is also introduced as a variant. The experiments are performed on ETTm1 and Electricity. The metric of variants is presented in relative value (’+’ indicates degraded performance, ’-’ indicates improved performance).}\vspace{-1mm}
\scalebox{0.6}{
\begin{tabular}{c|c|cccccccccccccccccc}
\toprule
\multicolumn{2}{c|}{Methods}&\multicolumn{2}{c|}{FilM}&\multicolumn{2}{c|}{LPU+MLP}&\multicolumn{2}{c|}{LPU+LSTM}&\multicolumn{2}{c|}{LPU+CNN}&\multicolumn{2}{c|}{LPU+attention}&\multicolumn{2}{c}{S4}\\
\midrule
\multicolumn{2}{c|}{Metric} & MSE  & MAE & MSE & MAE& MSE  & MAE& MSE  & MAE& MSE  & MAE& MSE  & MAE&\\
\midrule

\multirow{4}{*}{\rotatebox{90}{ETTm1}} 

&96  &\textbf{0.029} &\textbf{0.127}  &+0.0\%   & +0.0\%   & +12.1\% & +7.1\%    & +13.5\%   & +9.5\%   & +1.7\%  & +1.6\% &-- & -- \\
& 192  &\textbf{0.041} &\textbf{0.153}  &-1.5\% & -0.6\%  & +12.2\% & +8.5\%  & +10.8\%   & +7.8\%    & +2.0\%  & +3.3\% &-- & -- \\
& 336 & \textbf{0.053} & \textbf{0.175} &-1.7\% & -1.7\%  & +4.5\% & +4.0\%    & +10.8\%    & +6.3\%    & +4.5\% & +2.9\% &-- & --  \\
& 720 &\textbf{0.071}& \textbf{0.205}   &-0.9\%  & -1.0\%  & +5.5\% & +3.4\%  & +8.6\%    & +4.9\%    & +13.8\% & +7.8\% &-- & --  \\

\midrule
\multirow{4}{*}{\rotatebox{90}{Electricity}} 

&96  &\textbf{0.154}  &\textbf{0.247}  &+155\%  & +81\%   & +160\% & +84\%    & +330\%  & +155\%  & +242\% & +119\% & 128\% & 83\% \\
& 192 &\textbf{0.166} & \textbf{0.258} &+59\% & +39\%   & +121\%  & +67\%   & +224\%  & +117\%  & +264\% & +131\% & 124\% & 76\% \\
& 336 & \textbf{0.188}& \textbf{0.283} &+55\%  & +35\%   & +150\%  & +74\%   & +128\%  & +71\% & +183\% & +95\% & 117\% & 69\% \\
& 720 & \textbf{0.249}&\textbf{0.341}  &+33\%  & +25\%   & +154\%  & +73\%   & +192\%  & +95\%  & +312\% & +138\% & 89\% & 51\% \\ 
                        
\bottomrule
\end{tabular}
\label{tab:ablation_FEL}
}

\vskip -0.05in
\end{table*}
To prove the effectiveness of the FEL, we replace the FEL with multiple variants (MLP, LSTM, CNN, and Transformer). S4 is also introduced as a variant since it has a version with Legendre projection memory. The experimental results are summarized in Table \ref{tab:ablation_FEL}. FEL achieves the best performance compared with its LSTN, CNN, and Attention counterpart. MLP has comparable performance when input length is 192, 336, and 720. However, MLP suffers from insupportable memory usage, which is $N^2 L$ (while FEL is $N^2$). S4 achieves similar results as our LPU+MLP variant. Among all, the LPU+CNN variant shows the poorest performance.

\paragraph{Ablation of Multiscale and Data Normalization (RevIN)}

\begin{table*}[h]
\vskip -0.15in
\centering
\caption{Ablation studies of Normalization and Multiscale. Multiscale use 3 branches with: $T$, $2T$, and $4T$ as input sequence length. T is the predicted sequence length}\vspace{-1mm}
\scalebox{0.7}{
\begin{tabular}{c|c|cccccccccccccccccc}
\toprule
\multicolumn{2}{c|}{Dataset}&\multicolumn{2}{c|}{ETTm2}&\multicolumn{2}{c|}{Electricity}&\multicolumn{2}{c|}{Exchange}&\multicolumn{2}{c|}{Traffic}&\multicolumn{2}{c|}{Weather}&\multicolumn{2}{c}{Illness}&\multicolumn{2}{c}{}\\
\midrule
\multicolumn{2}{c|}{Metric} & MSE  & Relative & MSE & Relative & MSE & Relative & MSE & Relative & MSE & Relative & MSE & Relative &\\
\midrule

\multirow{4}{*}{\rotatebox{90}{Methods}} 

&Original  &0.301 & +16\% & 0.195 & +4.8\% & 0.534 & + 60\% & 0.528 & +19\% & 0.264 & +4.9\% & 3.55 & + 80\% \\

& Normalization & 0.268 & +3.6\% & 0.199 & +6.4\% & 0.456 & +36\% & 0.656 & +48\% & 0.256 & +1.5\% & 3.36 & +70\% \\	
& Multiscale & 0.259 & + 0.19\% &	0.187 &   +0\%	& 0.335	 & 0\% & 0.541 & +22\% & 0.253 & +0.50\% &	2.41 & +22\% \\
& With both & 0.271	& +4.7\% &	0.189	& +1.5\% & 0.398 & +19\% & 0.442 & +0\% &  0.253 & +0.4\% & 1.97 & +0\% \\



\bottomrule
\end{tabular}
\label{tab:ablation_normalize}
}
\vskip -0.05in
\end{table*}
The multiscale module leads to significant improvement on all datasets consistently. However, the data normalization achieves mixed performance, leading to improved performance on Traffic and Illness but a slight improvement on the rest. The Ablation study of RevIN data normalization and the mixture of multiscale experts employed is shown in Table \ref{tab:ablation_normalize}. 

\subsection{Other Studies}
Other supplemental studies are provided in Appendix. 1. A parameter sensitivity experiment (in Appendix \ref{app:parameter}) is carried out to discuss the choice of hyperparameter for M and N, where M is the frequency mode number and N is the Legendre Polynomials number. 
2. A noise injection experiment (in Appendix \ref{app:noise}) is conducted to show the robustness of our model. 3. A Kolmogorov-Smirnov (KS) test (in Appendix \ref{app:Distribution}) is conducted to discuss the similarity between the output distribution and the input. Our proposed FiLM has the best results on KS test, which supports our motivation in model design. 4. At last, (in Appendix \ref{app:time}), though failing short in training speed compared to some MLP based models like N-HiTS\cite{challu2022n}, FiLM is an efficient model with shorter per-step training time and smaller memory usage compared to baseline models in univariate forecasting tasks. However, it is worth mentioning that the training time will considerably prolong for multivariate forecasting tasks with many target variables. 

\section{Discussions and Conclusion}
\vspace{-2mm}
In long-term forecasting, the critical challenge is 
the trade-off between historical information preservation and noise reduction for accurate and robust forecasting. To address this challenge, we propose a Frequency improved Legendre Memory model, or {\bf FiLM}, to preserve historical information accurately and remove noisy signals. Moreover, we theoretically and empirically prove the effectiveness of the Legendre and Fourier projection employed in out model. Extensive experiments show that the proposed model achieves SOTA 
accuracy by a significant margin on six benchmark datasets. In particular, we would like to point out that our proposed framework is rather general and can be used as the building block for long-term forecasting in future research. It can be modified for different scenarios. For example, the Legendre Projection Unit can be replaced with other orthogonal functions such as Fourier, Wavelets, Laguerre Polynomials, Chebyshev Polynomials, etc. In addition, based on the properties of noises, Fourier Enhanced Layer is proved to be one of the best candidate in the framework. We plan to investigate more variants 
of this framework in the future.




\newpage
\bibliography{6_mybib}
\bibliographystyle{neurips_2022}

\newpage
\appendix

\section{Related Work}
\label{App:related_works}
In this section, we will give an overview of the related literature in time series forecasting. 



\paragraph{Traditional Time Series Models} The first generation of well-discussed time series model is the autoregressive family. ARIMA~\cite{box_arima2,box_distribution_1970} follows the Markov process and build recursive sequential forecasting. However, a plain autoregressive process has difficulty in dealing non-stationary sequences. Thus, ARIMA employed a pre-process iteration by differencing, which transforms the series to stationary. Still, ARIMA and related models have the linear assumption in the autoregressive process, which limits their usages in complex forecasting tasks. 

\paragraph{Deep Neural Network in Forecasting}
With the bloom of deep neural networks, recurrent neural networks (RNN) was designed for tasks involving sequential data. However, canonical RNN tends to suffer from gradient vanishing/exploding problem with long input because of its recurrent structure. Among the family of RNNs, LSTM~\cite{hochreiter_long_1997_lstm} and GRU~\cite{GRU_cho_et_al_2014} proposed gated structure to control the information flow to deal with the gradient vanishing/exploration problem. Although recurrent networks enjoy fast inference, they are slow to train and not parallelizable. Temporal convolutional network (TCN)~\cite{Think_globally_act_locally_tcn_time_series_2019} is another family for sequential tasks. However, limited to the reception field of the kernel, the long-term dependencies are hard to capture. Convolution is a parallelizable operation but expensive in inference.

\paragraph{Transformers}
With the innovation of Transformers in natural language processing~\cite{attention_is_all_you_need,Bert/NAACL/Jacob} and in computer vision tasks~\cite{Transformers-for-image-at-scale/iclr/DosovitskiyB0WZ21,DBLP:Global-filter-FNO-in-cv}, they are also discussed, renovated, and applied in time series forecasting~\cite{wen2022transformers}, especially the main attention module. Some works use temporal attention~\cite{dual-state-attention-rnn-qin} to capture long-range dependencies for time series. Others use the backbone of Transformer. Transformers usually employ an encoder-decoder architecture, with the self-attention and cross-attention mechanisms serving as the core layers. \cite{Log-transformer-shiyang-2019} invents a logsparse attention module to deal with the memory bottleneck for Transformer model. \cite{DBLP:conf/iclr/KitaevKL20-reformer} uses locality-sensitive
hashing to replace the attention module for less time complexity. \cite{haoyietal-informer-2021} proposes a probability sparse attention mechanism to deal with long-term forecasting. \cite{Autoformer} designs a decomposition Transformer architecture with an Auto-Correlation mechanism as an alternative for attention module. \cite{liu2022pyraformer} designs a low-complexity Pyramidal Attention for the long-term time forecasting tasks. \cite{FedFormer} proposes two attention modules which operate in frequency domain using Fourier or wavelet transformation. 
\paragraph{Orthogonal Basis and Neural Network}
Orthogonal basis project arbitrary functions onto a certain space and thus enable the representation learning in another view. Orthogonal basis family is easy to be discretized and to serve as an plug-in operation in neural networks. Recent studies began to realize the efficiency and effectiveness of Orthogonal basis, including the polynomial family and others (Fourier basis \& Multiwavelet basis). Fourier basis is first introduced for acceleration due to Fast Fourier algorithm, for example, acceleration of computing convolution~\cite{Hippo} or Auto-correlation function~\cite{Autoformer}. Fourier basis also serves as a performance boosting block: Fourier with Recurrent structure~\cite{FourierRecurrentUnits}, Fourier with MLP~\cite{Fourier-Neural-Operator,FNet} and Fourier in Transformer~\cite{FedFormer}. Multiwavelet transform is a more local filter (compared with 
Fourier) and a frequency decomposer. Thus, neural networks which employ multiwavelet filter usually exhibit a hierarchical structure and treat different frequency in different tunnels, e.g., \cite{wang2018multilevel,Multiwavelet-based-Operator-Learning,FedFormer}. Orthogonal Polynomials are naturally good selections of orthogonal basis. Legendre Memory Unit (LMU)~\cite{LMU} uses Legendre Polynomials for an orthogonal projection of input signals for a memory strengthening with the backbone of LSTM. The projection process is mathematically derived from delayed linear transfer function. HiPPO~\cite{Hippo}, based on LMU, proposes a novel mechanism (Scaled Legendre), which involves the function's full history (LMU uses rolling windowed history). In the subsequent work of HiPPO, the authors propose S4 model~\cite{S4} and gives the first practice on time series forecasting tasks. However, LMU and HiPPO share the same backbone (LSTM), which may limit their performance. 






\section{Algorithm Implementation}
\label{app:Alg}

\begin{algorithm}[h]
	\caption{Frequency Enhanced Layer}
	\label{alg:code0}
	\definecolor{codeblue}{rgb}{0.25,0.5,0.5}
	\lstset{
		backgroundcolor=\color{white},
		basicstyle=\fontsize{7.2pt}{7.2pt}\ttfamily\selectfont,
		columns=fullflexible,
		breaklines=true,
		captionpos=b,
		commentstyle=\fontsize{7.2pt}{7.2pt}\color{codeblue},
		keywordstyle=\fontsize{7.2pt}{7.2pt},
	}
    
    \begin{lstlisting}[language=python]
    class Freq_enhanced_layer(nn.Module):
        def __init__(self, in_channels, out_channels, modes1, modes2, compression=0):
            super(Freq_enhanced_layer, self).__init__()
            self.in_channels = in_channels
            self.out_channels = out_channels
            self.modes1 = modes1  #Number of Fourier modes to multiply, at most floor(N/2) + 1
            self.modes2 = modes2
            self.compression = compression 
            self.scale = (1 / (in_channels * out_channels))
            self.weights1 = nn.Parameter(self.scale * torch.rand(in_channels, out_channels, self.modes1))
            if compression>0: ## Low-rank approximation 
                self.weights0 = nn.Parameter(self.scale * torch.rand(in_channels, self.compression, dtype=torch.cfloat))
                self.weights1 = nn.Parameter(self.scale * torch.rand(self.compression, self.compression, len(self.index), dtype=torch.cfloat))
                self.weights2 = nn.Parameter(self.scale * torch.rand(self.compression, out_channels, dtype=torch.cfloat))

        def forward(self, x):
            B, H,E, N = x.shape
            # Compute Fourier coefficients up to factor of e^(- something constant)
            x_ft = torch.fft.rfft(x)
            # Multiply relevant Fourier modes
            out_ft = torch.zeros(B, H, self.out_channels, x.size(-1)//2 + 1)
            if self.compression == 0:
                a = x_ft[:, :, :, :self.modes1]
                out_ft[:, :, :, :self.modes1] = torch.einsum("bjix,iox->bjox", a, self.weights1)
            else:
                a = x_ft[:, :, :, :self.modes2]
                a = torch.einsum("bjix,ih->bjhx", a, self.weights0)
                a = torch.einsum("bjhx,hkx->bjkx", a, self.weights1)
                out_ft[:, :, :, :self.modes2] = torch.einsum("bjkx,ko->bjox", a, self.weights2)
            # Return to physical space
            x = torch.fft.irfft(out_ft, n=x.size(-1))
            return x
    \end{lstlisting}
\end{algorithm}


\begin{algorithm}[h]
	\caption{LPU layer}
	\label{alg:code1}
	\definecolor{codeblue}{rgb}{0.25,0.5,0.5}
	\lstset{
		backgroundcolor=\color{white},
		basicstyle=\fontsize{7.2pt}{7.2pt}\ttfamily\selectfont,
		columns=fullflexible,
		breaklines=true,
		captionpos=b,
		commentstyle=\fontsize{7.2pt}{7.2pt}\color{codeblue},
		keywordstyle=\fontsize{7.2pt}{7.2pt},
	}
    
    \begin{lstlisting}[language=python]
from scipy import signal
from scipy import special as ss
class LPU(nn.Module):
    def __init__(self, N=256, dt=1.0, discretization='bilinear'):
        # N: the order of the Legendre projection 
        # dt: step size - can be roughly inverse to the length of the sequence
        super(LPU,self).__init__()
        self.N = N
        A,B = transition(N) ### LMU projection matrix
        A,B, _, _, _ = signal.cont2discrete((A, B, C, D), dt=dt, method=discretization) 
        B = B.squeeze(-1)
        self.register_buffer('A', torch.Tensor(A)) 
        self.register_buffer('B', torch.Tensor(B)) 
    def forward(self, inputs):  
        # inputs: (length, ...) 
        # output: (length, ..., N) where N is the order of the Legendre projection
        c = torch.zeros(inputs.shape[:-1] + tuple([self.N]))
        cs = []
        for f in inputs.permute([-1, 0, 1]):
            f = f.unsqueeze(-1)
            new = f @ self.B.unsqueeze(0) # [B, D, H, 256]
            c = F.linear(c, self.A) + new
            cs.append(c)
        return torch.stack(cs, dim=0)
    def reconstruct(self, c):
        a = (self.eval_matrix @ c.unsqueeze(-1)).squeeze(-1)
        return (self.eval_matrix @ c.unsqueeze(-1)).squeeze(-1)
    \end{lstlisting}
    
\end{algorithm}
\section{Dataset and Implementation Details}
\label{app:section_dataset_implem}
\subsection{Dataset Details}
\label{app:dataset}
In this subsection, we summarize the details of the datasets used in this paper as follows: 1) ETT~\cite{haoyietal-informer-2021} dataset contains two sub-dataset: ETT1 and ETT2, collected from two separated counties. Each of them has two versions of sampling resolutions (15min \& 1h). ETT dataset contains multiple time series of electrical loads and one time sequence of oil temperature. 2) Electricity\footnote{https://archive.ics.uci.edu/ml/datasets/ElectricityLoadDiagrams 20112014} dataset contains the electricity consumption for more than three hundred clients with each column corresponding to one client. 3) Exchange~\cite{lai2018modeling-exchange-dataset} dataset contains the current exchange of eight countries. 4) Traffic\footnote{http://pems.dot.ca.gov} dataset contains the occupation rate of freeway systems in California, USA. 5) Weather\footnote{https://www.bgc-jena.mpg.de/wetter/} dataset contains 21 meteorological indicators for a range of one year in Germany. 6) Illness\footnote{https://gis.cdc.gov/grasp/fluview/fluportaldashboard.html} dataset contains the influenza-like illness patients in the United States. Table \ref{tab:dataset} 
\begin{table}[t]
\caption{Details of benchmark datasets.}
\label{tab:dataset}
\vskip 0.15in
\begin{center}
\begin{small}
\begin{sc}
\begin{tabular}{l|cccr}
\toprule
Dataset & Length & Dimension & Frequency \\
\midrule
ETTm2 & 69680 & 8 & 15 min\\
Exchange & 7588 & 9 & 1 day\\
Weather & 52696 & 22 & 10 min & \\
Electricity & 26304 & 322 & 1h & \\
ILI & 966 & 8 & 7 days\\
Traffic & 17544 & 863 & 1h & \\
\bottomrule
\end{tabular}
\end{sc}
\end{small}
\end{center}
\vskip -0.1in
\end{table}
summarizes all the features for the six benchmark datasets. They are all split into the training set, validation set and test set by the ratio of 7:1:2 during modeling.

\subsection{Implementation Details}
\label{app:exp:implement}
Our model is trained using ADAM \cite{kingma_adam:_2017} optimizer with a learning rate of $1e^{-4}$ to  $1e^{-3}$. The batch size is set to 32 (It depends on the GPU memory used in the experiment. In fact, a batch size up to 256 does not deteriorate the performance but with faster training speed if larger memory GPU or multiple GPUs is used). The default training process is 15 epochs without any early stopping. We save the model with the lowest loss on the validation set for the final testing. The mean square error (MSE) and mean absolute error (MAE) are used as metrics. All experiments are repeated 5 times and the mean of the metrics is reported as the final results. All the deep learning networks are implemented using PyTorch \cite{NEURIPS2019_9015_pytorch} and trained on NVIDIA V100 32GB GPUs/NVIDIA V100 16GB GPUs.

\subsection{Experiment Error Bars}
We train our model 5 times and calculate the error bars for FiLM and SOTA model FEDformer to compare the robustness, which is summarized in Table~\ref{tab:std}. It can be seen that the overall performance of the proposed FiLM is better than that of the SOTA FEDformer model.

\label{app:error_bars}
\vspace{-.2cm}
\begin{table}[ht]
\centering
\caption{MSE with error bars (Mean and STD) for FiLM and FEDformer baseline for multivariate long-term forecasting. All experiments are repeated 5 times.}
\begin{small}
\scalebox{0.75}{
\begin{tabular}{c|c|ccccc}
\toprule
\multicolumn{2}{c|}{MSE}& ETTm2 & Electricity & Exchange & Traffic\\ 
\midrule
\multirow{4}{*}{\rotatebox{90}{FiLM}} 
& 96 & 0.165 $\pm$ 0.0051 & 0.153$\pm$ 0.0014 & 0.079$\pm$ 0.002 & 0.416$\pm$ 0.010 \\
& 192 & 0.222 $\pm$ 0.0038 & 0.165$\pm$ 0.0023 & 0.159$\pm$ 0.011 & 0.408$\pm$ 0.007 \\
& 336 & 0.277 $\pm$ 0.0021 & 0.186$\pm$ 0.0018 & 0.270$\pm$ 0.018 & 0.425$\pm$ 0.007 \\
& 720 & 0.371 $\pm$ 0.0066 & 0.236$\pm$ 0.0022 & 0.536$\pm$ 0.026 & 0.520$\pm$ 0.003 \\
\midrule
\multirow{4}{*}{\rotatebox{90}{FED-f}} 
& 96 & 0.203 $\pm$ 0.0042 & 0.194 $\pm$ 0.0008 & 0.148 $\pm$ 0.002 & 0.217 $\pm$ 0.008 \\
& 192 & 0.269 $\pm$ 0.0023 & 0.201$\pm$ 0.0015 & 0.270$\pm$ 0.008 & 0.604 $\pm$ 0.004 \\
& 336 & 0.325 $\pm$ 0.0015 & 0.215$\pm$ 0.0018 & 0.460$\pm$ 0.016 & 0.621 $\pm$ 0.006 \\
& 720 & 0.421 $\pm$ 0.0038 & 0.246$\pm$ 0.0020 & 1.195$\pm$ 0.026 & 0.626 $\pm$ 0.003 \\
\bottomrule
\end{tabular}
}
\label{tab:std}
\end{small}
\vskip -0.1in
\end{table}

\subsection{Univariate Forecasting Results}
\label{app:exp:uni_bench}
The univariate benchmark results are summarized in Table~\ref{tab:uni-benchmarks-large}.
\begin{table*}[h]
\centering
\caption{Univariate long-term forecasting results on six datasets with various input length and prediction horizon $O \in \{96,192,336,720\}$. A lower MSE indicates better performance. All experiments are repeated 5 times.}
\vskip 0.05in
\scalebox{0.60}{
\begin{tabular}{c|c|cccccccccccccc}
\toprule
\multicolumn{2}{c|}{Methods}&\multicolumn{2}{c|}{FiLM}&\multicolumn{2}{c|}{FEDformer}&\multicolumn{2}{c|}{Autoformer}&\multicolumn{2}{c|}{S4}&\multicolumn{2}{c|}{Informer}&\multicolumn{2}{c|}{LogTrans}&\multicolumn{2}{c}{Reformer}\\
\midrule
\multicolumn{2}{c|}{Metric} & MSE  & MAE & MSE & MAE& MSE  & MAE& MSE  & MAE& MSE  & MAE& MSE  & MAE& MSE  & MAE\\
\midrule
\multirow{4}{*}{\rotatebox{90}{$ETTm2$}} 
& 96 & 0.065 &\textbf{0.189}  &\textbf{0.063} &0.189 & 0.065 & 0.189 &0.153 &0.318 & 0.088 & 0.225 & 0.075 & 0.208 & 0.076  &0.214   \\
& 192 & \textbf{0.094} & \textbf{0.233} &0.102 &0.245 & 0.118 & 0.256 &0.183 &0.350 & 0.132 &0.283 & 0.129 &0.275  & 0.132  & 0.290   \\
& 336 & \textbf{0.124} & \textbf{0.274} &0.130 &0.279 & 0.154 & 0.305 &0.204 & 0.367 &0.180 &0.336 & 0.154 &0.302  & 0.160  & 0.312    \\
& 720 & \textbf{0.173} & \textbf{0.323} &0.178 &0.325 & 0.182 &0.335 &0.482 &0.567  &0.300  &0.435 & 0.160 &0.321  & 0.168  &0.335     \\
\midrule
\multirow{4}{*}{\rotatebox{90}{$Electricity$}} 
& 96  & \textbf{0.154} & \textbf{0.247} &0.253 &0.370  &0.341 &0.438 &0.351 &0.452 & 0.484& 0.538 &0.288& 0.393& 0.274 & 0.379   \\
& 192 & \textbf{0.166} & \textbf{0.258} &0.282 &0.386 &0.345 &0.428 &0.373 &0.455 & 0.557& 0.558 &0.432 & 0.483 & 0.304 &0.402   \\
& 336 & \textbf{0.188} & \textbf{0.283} &0.346 &0.431 &0.406 &0.470 &0.408 &0.477 & 0.636& 0.613 &0.430 &0.483 & 0.370 &0.448  \\
& 720 & \textbf{0.249} & \textbf{0.341} &0.422 &0.484 &0.565 &0.581 &0.472 &0.517 & 0.819& 0.682 &0.491 &0.531 & 0.460 &0.511   \\
\midrule

\multirow{4}{*}{\rotatebox{90}{$Exchange$}} 
&96  & \textbf{0.110} & \textbf{0.259} &0.131 &0.284 & 0.241 & 0.387 &0.344 &0.482  & 0.591  &0.615  &0.237 &0.377  &0.298  &0.444    \\
& 192 & \textbf{0.207} & \textbf{0.352} &0.277 &0.420 &0.300  &0.369 &0.362 &0.494  &1.183   &0.912 &0.738  &0.619  &0.777  & 0.719   \\
& 336 & \textbf{0.327} & \textbf{0.461} &0.426 &0.511 &0.509  &0.524 &0.499 &0.594  &1.367  &0.984  &2.018  &1.070  &1.832  &1.128     \\
& 720 & \textbf{0.811} & \textbf{0.708} &1.162 &0.832 &1.260  &0.867 &0.552 & 0.614 &1.872  &1.072  &2.405  &1.175  &1.203  &0.956     \\
\midrule
\multirow{4}{*}{\rotatebox{90}{$Traffic$}} 
&96  & \textbf{0.144} & \textbf{0.215} &0.170 &0.263  &0.246 &0.346 &0.194 &0.290 & 0.257 & 0.353 &0.226 &0.317 &  0.313 & 0.383\\
& 192 & \textbf{0.120} & \textbf{0.199} &0.173 &0.265 &0.266 &0.370 &0.172 &0.272 & 0.299 & 0.376 &0.314 &0.408 & 0.386 & 0.453\\
& 336 & \textbf{0.128} & \textbf{0.212} &0.178 &0.266 &0.263 &0.371 &0.178 &0.278 & 0.312 & 0.387 &0.387 &0.453 & 0.423 & 0.468\\
& 720 & \textbf{0.153} & \textbf{0.252} &0.187 &0.286 &0.269 &0.372 &0.263 &0.386 & 0.366 & 0.436 &0.491 &0.437 & 0.378 & 0.433\\
\midrule
\multirow{4}{*}{\rotatebox{90}{$Weather$}}
& 96  & \textbf{0.0012} & \textbf{0.026} & 0.0035 & 0.046 & 0.011 &0.081 &0.0061 &0.065 & 0.0038 & 0.044 & 0.0046 &0.052 & 0.012 & 0.087 \\
& 192 & \textbf{0.0014}& \textbf{0.029}&0.0054 &0.059 &0.0075 &0.067 &0.0067 &0.067 & 0.0023 & 0.040 &0.0056 & 0.060 & 0.0098 & 0.079  \\
& 336 & \textbf{0.0015} & \textbf{0.030} &0.0041 &0.050 &0.0063 &0.062 &0.0025 &0.0381 & 0.0041 & 0.049 &0.0060 &0.054& 0.0050 & 0.059 \\
& 720 & \textbf{0.0022} & \textbf{0.037} &0.015 &0.091 &0.0085 &0.070 &0.0074 &0.0736 & 0.0031 & 0.042 &0.0071 &0.063& 0.0041 & 0.049 \\
\midrule
\multirow{4}{*}{\rotatebox{90}{$ILI$}} 
& 24 & \textbf{0.629} & \textbf{0.538} &0.693 &0.629 &0.948 &0.732 &0.866 &0.584 & 5.282 &2.050 &3.607 & 1.662 & 3.838 & 1.720  \\
& 36 & \textbf{0.444} & \textbf{0.481} &0.554 &0.604 &0.634 &0.650 &0.622 &0.532 & 4.554 &1.916 &2.407 & 1.363 & 2.934 & 1.520 \\
& 48 & \textbf{0.557} & \textbf{0.584} &0.699 &0.696 &0.791 &0.752 &0.813 &0.679 & 4.273 &1.846 &3.106 & 1.575 & 3.754 & 1.749\\
& 60 & \textbf{0.641} & \textbf{0.644} &0.828 &0.770 &0.874 &0.797 &0.931 &0.747 & 5.214 &2.057 &3.698 & 1.733 & 4.162 & 1.847\\
\bottomrule
\end{tabular}
}
\label{tab:uni-benchmarks-large}
\vskip -0.10in
\end{table*}

\subsection{ETT Full Benchmark}
\label{app:exp:ett_benchmark}
We present the full-benchmark on four ETT datasets \cite{haoyietal-informer-2021} in Table \ref{tab:multi-benchmarks-ett} (multivariate forecasting) and Table \ref{tab:uni-benchmarks-ett} (univariate forecasting). The ETTh1 and ETTh2 are recorded hourly while ETTm1 and ETTm2 are recorded every 15 minutes. The time series in ETTh1 and ETTm1 follow the same pattern, and the only difference is the sampling rate, similarly for ETTh2 and ETTm2. On average, our 
FiLM yields a \textbf{14.0\%} relative MSE reduction for multivariate forecasting, and a \textbf{16.8\%} reduction for univariate forecasting over the SOTA results from FEDformer.
\label{app:ETT}

\begin{table*}[h]
\centering
\caption{Multivariate long-term forecasting results on ETT full benchmark. The best results are highlighted in bold. A lower MSE indicates better performance. All experiments are repeated 5 times.}
\vskip 0.05in
\scalebox{0.75}{
\begin{tabular}{c|c|cccccccccccccccccc}
\toprule
\multicolumn{2}{c|}{Methods}&\multicolumn{2}{c|}{FiLM}&\multicolumn{2}{c|}{FEDformer}&\multicolumn{2}{c|}{Autoformer}&\multicolumn{2}{c|}{S4}&\multicolumn{2}{c|}{Informer}&\multicolumn{2}{c|}{LogTrans}&\multicolumn{2}{c}{Reformer}\\
\midrule
\multicolumn{2}{c|}{Metric} & MSE  & MAE & MSE & MAE& MSE  & MAE& MSE  & MAE& MSE  & MAE& MSE  & MAE& MSE  & MAE\\
\midrule
\multirow{4}{*}{\rotatebox{90}{$ETTh1$}}
& 96 &\textbf{0.371} &\textbf{0.394} &0.376 &0.419  & 0.449& 0.459 &0.949 &0.777 & 0.865 & 0.713& 0.878& 0.740& 0.837& 0.728\\
& 192 &\textbf{0.414} &\textbf{0.423} &0.420 &0.448  & 0.500& 0.482 &0.882 &0.745 & 1.008 & 0.792& 1.037& 0.824& 0.923& 0.766\\
& 336 &\textbf{0.442} &\textbf{0.445} &0.459 &0.465  & 0.521& 0.496 &0.965 &0.75 & 1.107 & 0.809& 1.238& 0.932& 1.097& 0.835\\
& 720 &\textbf{0.465} &\textbf{0.472} &0.506 &0.507  & 0.514& 0.512 &1.074&0.814 & 1.181 &0.865&  1.135& 0.852& 1.257& 0.889\\
\midrule
\multirow{4}{*}{\rotatebox{90}{$ETTh2$}}
& 96  &\textbf{0.284} &\textbf{0.348} &0.346 &0.388 & 0.358& 0.397 &1.551&	0.968 & 3.755& 1.525& 2.116& 1.197 &2.626 &1.317\\
& 192 &\textbf{0.357} &\textbf{0.400} &0.429 &0.439 & 0.456& 0.452 &2.336	 &1.229 & 5.602& 1.931& 4.315& 1.635 &11.12 &2.979\\
& 336 &\textbf{0.377} &\textbf{0.417} &0.482 &0.480 & 0.482& 0.486 &2.801	 &1.259 & 4.721& 1.835& 1.124& 1.604 &9.323 &2.769\\
& 720 &\textbf{0.439} &\textbf{0.456} & 0.463 & 0.474 & 0.515& 0.511 &2.973	 &1.333 & 3.647& 1.625& 3.188& 1.540 &3.874 &1.697\\
\midrule
\multirow{4}{*}{\rotatebox{90}{$ETTm1$}}
& 96  &\textbf{0.302} &\textbf{0.345} &0.378 &0.418 & 0.505& 0.475 &0.640	 &0.584 & 0.672& 0.571& 0.600& 0.546 &0.538 &0.528\\
& 192 &\textbf{0.338} &\textbf{0.368} & 0.426 & 0.441 & 0.553& 0.496 &0.570	 &0.555 & 0.795& 0.669& 0.837& 0.700 &0.658 &0.592\\
& 336 &\textbf{0.373} &\textbf{0.388} & 0.445 & 0.459 & 0.621& 0.537 &0.795& 0.691& 1.212& 0.871& 1.124& 0.832 &0.898 &0.721\\
& 720 &\textbf{0.420} &\textbf{0.420} & 0.543 & 0.490 & 0.671& 0.561 &0.738 &0.655 & 1.166& 0.823& 1.153& 0.820 &1.102 &0.841\\
\midrule
\multirow{4}{*}{\rotatebox{90}{$ETTm2$}} &96 &\textbf{0.165} &\textbf{0.256} & 0.203 & 0.287 &0.255  &0.339  &0.705	 &0.690  &0.365  &0.453  &0.768  &0.642  &0.658  &0.619 \\
                        & 192 &\textbf{0.222} &\textbf{0.296} & 0.269 & 0.328 &0.281 &0.340  &0.924	 &0.692 &0.533  &0.563  &0.989  &0.757  &1.078  &0.827 \\
                        & 336 &\textbf{0.277} &\textbf{0.333} & 0.325 & 0.366  &0.339  &0.372  &1.364	 & 0.877 &1.363&0.887  &1.334  &0.872  &1.549  &0.972 \\
                        & 720 &\textbf{0.371} &\textbf{0.389} & 0.421 & 0.415  &0.422  &0.419  &2.074 &1.074  &3.379  &1.338 & 3.048 &1.328  &2.631  &1.242 \\
\bottomrule
\end{tabular}
}
\label{tab:multi-benchmarks-ett}
\end{table*}

\begin{table*}[h]
\centering
\caption{Univariate long-term forecasting results on ETT full benchmark. The best results are highlighted in bold. A lower MSE indicates better performance. All experiments are repeated 5 times.}
\vskip 0.05in
\scalebox{0.75}{
\begin{tabular}{c|c|cccccccccccccccccc}
\toprule
\multicolumn{2}{c|}{Methods}&\multicolumn{2}{c|}{FiLM}&\multicolumn{2}{c|}{FEDformer}&\multicolumn{2}{c|}{Autoformer}&\multicolumn{2}{c|}{S4}&\multicolumn{2}{c|}{Informer}&\multicolumn{2}{c|}{LogTrans}&\multicolumn{2}{c}{Reformer}\\
\midrule
\multicolumn{2}{c|}{Metric} & MSE  & MAE & MSE & MAE& MSE  & MAE& MSE  & MAE& MSE  & MAE& MSE  & MAE& MSE  & MAE\\
\midrule

\multirow{4}{*}{\rotatebox{90}{$ETTh1$}}
&96 &\textbf{0.055}&\textbf{0.178}&0.079 &0.215 &0.071 &0.206 &0.316 &0.490 &	0.193&	0.377&	0.283&	0.468&	0.532&	0.569\\
&192 &\textbf{0.072}&\textbf{0.207}&0.104 &0.245 & 0.114&	0.262 & 0.345&0.516 &	0.217&	0.395&	0.234&	0.409&	0.568&	0.575\\
&336 &\textbf{0.083}&\textbf{0.229}&0.119 &0.270 &0.107&0.258 &0.825 &0.846 &	0.202&	0.381&	0.386&	0.546&	0.635&	0.589\\
&720 &\textbf{0.090}&\textbf{0.240}&0.127 &0.280 &0.126&0.283 &0.190 &0.355 &	0.183&	0.355&	0.475&	0.628&	0.762&	0.666\\
\midrule
\multirow{4}{*}{\rotatebox{90}{$ETTh2$}}
&96 &\textbf{0.127}&0.272&0.128 &\textbf{0.271} & 0.153&	0.306 & 0.381&0.501 &	0.213&	0.373&	0.217&	0.379&	1.411&	0.838\\
&192 &\textbf{0.182}&0.335&0.185 &\textbf{0.330} & 0.204&	0.351 & 0.332&0.458 &	0.227&	0.387&	0.281&	0.429&	5.658&	1.671\\
&336 &\textbf{0.204}&\textbf{0.367}&0.231 &0.378 & 0.246&	0.389 & 0.655&0.670 &	0.242&	0.401&	0.293&	0.437&	4.777&	1.582\\
&720 &\textbf{0.241}&\textbf{0.396}&0.278 &0.420 & 0.268&	0.409 & 0.630&0.662 &	0.291&	0.439&	0.218&	0.387&	2.042&	1.039\\
\midrule
\multirow{4}{*}{\rotatebox{90}{$ETTm1$}}
&96 &\textbf{0.029}&\textbf{0.127}&0.033 &0.140 & 0.056&	0.183 & 0.651&0.733 &	0.109&	0.277&	0.049&	0.171&	0.296&	0.355\\
&192 &\textbf{0.041}&\textbf{0.153}&0.058 &0.186 & 0.081&	0.216 & 0.190& 0.372&	0.151&	0.310&	0.157&	0.317&	0.429&	0.474\\
&336 &\textbf{0.053}&\textbf{0.175}&0.071 &0.209 & 0.076&	0.218 & 0.428&0.581 &	0.427&	0.591&	0.289&	0.459&	0.585&	0.583\\
&720 &\textbf{0.071}&\textbf{0.205}&0.102 &0.250 & 0.110&	0.267 & 0.254&0.433 &	0.438&	0.586&	0.430&	0.579&	0.782&	0.730\\
\midrule
\multirow{4}{*}{\rotatebox{90}{$ETTm2$}} 
& 96 &0.065&\textbf{0.189}&\textbf{0.063} &0.189 & 0.065 & 0.189  & 0.153&0.318 & 0.088 & 0.225 & 0.075 & 0.208 & 0.076  &0.214   \\
& 192 &\textbf{0.094}&\textbf{0.233}& 0.102 & 0.245 & 0.118 & 0.256  &0.183 &0.350 & 0.132 &0.283 & 0.129 &0.275  & 0.132  & 0.290   \\
& 336 &\textbf{0.124}&\textbf{0.274}& 0.130 & 0.279 & 0.154 & 0.305  &0.204 &0.367  &0.180 &0.336 & 0.154 &0.302  & 0.160  & 0.312    \\
& 720 &\textbf{0.173}&\textbf{0.323}& 0.178 & 0.325 & 0.182 &0.335  &0.482 &0.567  &0.300  &0.435 & 0.160 &0.321  & 0.168  &0.335     \\
\bottomrule
\end{tabular}
}
\label{tab:uni-benchmarks-ett}
\end{table*}

\section{Low-rank Approximation for FEL}
\label{app:LRA}
With the low-rank approximation of learnable matrix in Fourier Enhanced Layer significantly reducing our parameter size, here we study its effect on model accuracy on two typical datasets as shown in Table~\ref{tab:FEL_LRA}.
\begin{table*}[h]

\centering
\caption{Low-rank Approximation (LRA) study for frequency enhanced layer: Comp. K=0 means default version without LRA, 1 means the largest compression using K=1.}
\scalebox{0.75}{
\begin{tabular}{c|c|cccccccccc}
\toprule

\multicolumn{2}{c|}{Comp. K}&\multicolumn{2}{c|}{0}&\multicolumn{2}{c|}{16}&\multicolumn{2}{c|}{4}&\multicolumn{2}{c}{1}\\

\midrule
\multicolumn{2}{c|}{Metric} & MSE  & MAE & MSE & MAE& MSE  & MAE& MSE  & MAE \\
\midrule


\multirow{4}{*}{\rotatebox{90}{$ETTh1$}} 

&96  &\textbf{0.371} &\textbf{0.394}    &0.371  &0.397  &0.373  &0.399   & 0.391   & 0.418    \\
& 192  &0.414 &\textbf{0.423}  &0.414  &0.425  &\textbf{0.413}  &0.426   &  0.437  &  0.445    \\
& 336 & \textbf{0.442} & 0.445 &0.452  &0.451  &0.445  &\textbf{0.444}   &0.460    &0.458       \\
& 720 &\textbf{0.454}& \textbf{0.451}   &0.460  &0.472  &0.461  &0.471   &0.464    &0.476      \\

\midrule

\multirow{4}{*}{\rotatebox{90}{$Weather$}} 

&96  &\textbf{0.199} &\textbf{0.262}    &0.200  &0.266  &0.199  &0.263   &0.198    &0.261      \\
& 192  &0.228 &0.288  &0.232  &0.298  &0.227  &0.287   &\textbf{0.226}    &\textbf{0.285}       \\
& 336 & 0.267 & 0.323 &0.266  &0.320  &\textbf{0.253}  &\textbf{0.314}   &0.264    &0.316       \\
& 720 &0.319& 0.361   &\textbf{0.314}  &\textbf{0.352}  &0.319  &0.361   &0.314    &0.354      \\

\midrule
\multicolumn{2}{c|}{Parameter size}&\multicolumn{2}{c|}{100\%}&\multicolumn{2}{c|}{6.4\%}&\multicolumn{2}{c|}{1.6\%}&\multicolumn{2}{c}{0.4\%}\\
\bottomrule
\end{tabular}
\label{tab:FEL_LRA}
}
\end{table*}




\section{Theoretical Analysis}

\subsection{Theorem 1}
The proof is a simple extension of the Proposition 6 in \cite{Hippo}. We omit it for brevity.

\subsection{Theorem 2}
As we have $x_t = Ax_{t-1} + b - \epsilon_{t-1}$ for $t=2,3,...\theta$, we recursively use them and the following result holds:
\begin{align}
    x_{t} &= Ax_{t-1} +  b + \epsilon_{t-1}\notag\\
    &= A(Ax_{t-2} +  b + \epsilon_{t-2})+  b + \epsilon_{t-1}\notag\\
    &=A^2x_{t-2} + Ab + b + A\epsilon_{t-2} + \epsilon_{t-1}\notag\\
    &\cdots\notag\\
    &=A^{\theta}x_{t-\theta} + \sum_{i=1}^{\theta-1}A^ib+ \underbrace{\sum_{i=1}^{\theta-1}A^i\epsilon_{t-i}}_{(*)}\notag.
\end{align}
Following Hoeffding inequality, for $\mu>0$ we have
\begin{align}
    \mathbbm{P}\left(|(*)|\ge \mu\right)\le \exp\left(-\frac{2\mu^2}{\sum_{i=1}^{\theta -1}\|A^i\epsilon_{t-1}\|_{\psi_2}^2}\right),
\end{align}
where $\|\cdot\|_{\psi_2}$ is the Orlicz norm defined as
\begin{align}
    \|X\|_{\psi_2} :=\inf\left\{c\ge 0: \mathbbm{E}[\exp(X^2/c^2)]\le 2\right\}\notag.
\end{align}
Since we require $A$ being unitary, we will have $\|A\epsilon\|_2^2 = \|\epsilon\|_2^2$ and it implies $\|A^i\epsilon_{t-1}\|_{\psi_2}^2\| = \|\epsilon_{t-1}\|_{\psi_2}^2 =   \mathcal{O}(\sigma^2)$ for $i=1,2,...,\theta$. The desirable result follows by setting $\mu = \mathcal{O}(\sqrt{\theta}\sigma)$.

\subsection{Theorem 3}


As we keep first $s$ columns selected, $P(A)-A$ has all $0$ elements in first $s$ columns. We thus ignore them and consider the approximation quality on $\tilde{A}\in\mathbb{R}^{d\times (n-s)}$ with the sampled columns.  Via the similar analysis in Appendix C of \cite{FedFormer}, with high probability we have $\| \tilde{A}-P(A) \|_F \leq (1+\epsilon) \|\tilde{A} - \tilde{A}_k\|_F$, where $\tilde{A}_k$ is the ``best” rank-$k$ approximation provided by truncating the singular value decomposition (SVD) of $\tilde{A}$, and where $\|\cdot\|_F$
is the Frobenius norm. As we assume the element in last $n - s$ columns of $A$ is smaller than $a_{\min}$, one can verify  $\| \tilde{A}-P(A) \|_F\le \sqrt{d \times (n-s)}a_{\min}$ and desirable result follows immediately.  

\section{Parameter Sensitivity}
\label{app:parameter}
\paragraph{Influence of Legendre Polynomial number $N$ and Frequency mode number $M$}
\begin{figure*}[h]
    \centering
    \includegraphics[width=0.8\linewidth]{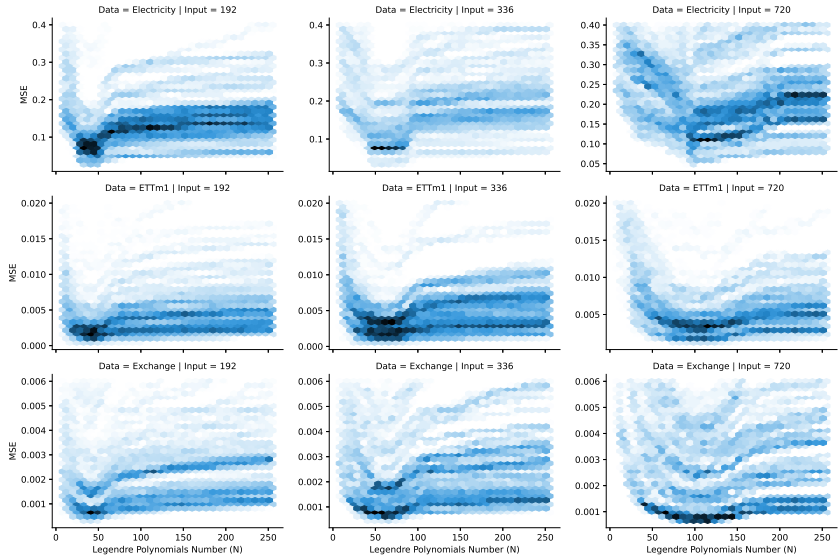}
    \caption{The reconstruction error (MSE) vs. Legendre Polynomial number ($N$) on three datasets with three different input lengths.}
    \label{fig:MSE_vs_Hippo_N}
\end{figure*}
The experimental results on three different datasets (ETTm1, Electricity, and Exchange) in Figure \ref{fig:MSE_vs_Hippo_N} show the optimal choice of Legendre Polynomials number ($N$) when we aim to minimize the reconstruction error (in MSE) on the historical data. The MSE error decreases sharply at first and saturates at an optimal point, where $N$ is in proportion to the input length. For input sequence with lengths of 192, 336, and 720, $N$ $\approx$ 40, 60, and 100 gives the minimal MSE,  respectively. 

\begin{figure*}[h]
    \centering
    \includegraphics[width=0.99\linewidth]{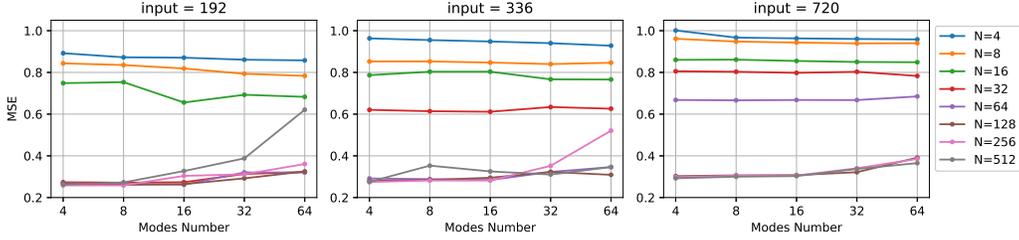}
    \caption{The MSE error of univariate time series forecasting task on Electricity dataset with different Legendre Polynomials number ($N$), mode number and input length. Left: input length = 192. Mid: input length = 336. Right: input length = 720.}
    \label{fig:MSE_vs_modes_vs_N}
\end{figure*}

Figure \ref{fig:MSE_vs_modes_vs_N} shows the MSE error of time series forecasting on Electricity dataset, with different Legendre Polynomials number ($N$), mode number and input length. We observe that, when enlarging $N$, the model performance saturates at an optimal point. For example, in Figure \ref{fig:MSE_vs_modes_vs_N} Left (input length=192), the best performance is reached when $N>64$. While in Figure \ref{fig:MSE_vs_modes_vs_N} Right (input length=720), the best performance is reached when $N$ is larger than 128. Another influential parameter is the mode number. From Figure \ref{fig:MSE_vs_modes_vs_N} we observe that a small mode number will lead to better performance, as module with small mode number works as a denoising filter.
\section{Noise Injection Experiment}
\label{app:noise}
Our model's robustness in long-term forecasting tasks can be demonstrated using a series of noise injection experiments as shown in Table \ref{tab:ablation_noise}. As we can see, adding Gaussian noise in the training/test stage has limited effect on our model's performance, since the 
deterioration is less than 1.5\% in the worst case. The model's robustness is consistent across various forecasting horizons. Note that adding the noise in testing stage other than the training stage will even improve our performance by 0.4\%, which further supports our claim of robustness. 
\begin{table*}[h]
\centering
\caption{Noise injection studies. A 0.3*$\mathcal{N}\left(0,1\right)$ Gaussian noise is introduced into our training/testing. We conduct 4 sets of experiments with/without noise in training and test phases. The experiments are performed on ETTm1 and Electricity with different output lengths. The metric of variants is presented in relative value ('+' indicates degraded performance, and '-' indicates improved performance).}
\scalebox{0.75}{
\begin{tabular}{c|c|cccccccc}
\toprule

\multicolumn{2}{c|}{Training}&\multicolumn{4}{c|}{noise}&\multicolumn{4}{c}{with noise}\\
\midrule
\multicolumn{2}{c|}{Testing}&\multicolumn{2}{c|}{without noise}&\multicolumn{2}{c|}{with noise}&\multicolumn{2}{c|}{without noise}&\multicolumn{2}{c}{with noise}\\

\midrule
\multicolumn{2}{c|}{Metric} & MSE  & MAE & MSE & MAE& MSE  & MAE& MSE  & MAE \\
\midrule
\multirow{4}{*}{\rotatebox{90}{$ETTh1$}} 

&96  &\textbf{0.371} &\textbf{0.394}    &-1.6\%  & -2.0\% & -0.0\% & -0.0\%  & -1.6\%   & -2.0\%     \\
& 192  &\textbf{0.414} &\textbf{0.423}  &-0.5\%  & -1.4\%    & +0.5\% & -0.0\%  & -0.5\%   & -1.4\%    \\
& 336 & \textbf{0.442} & \textbf{0.445} &-1.8\% & -0.9\%    & -0.9\% & +1.3\%  & -3.2\%   & -1.6\%     \\
& 720 &\textbf{0.465}& \textbf{0.472}   &+0.2\%     & -0.2\%     &+0.9\%    & +0.9\%  & -0.6\%    & -0.4\%    \\


\bottomrule
\end{tabular}
\label{tab:ablation_noise}
}
\end{table*}
\section{Distribution Analysis of Forecasting Output}
\label{app:Distribution}
\subsection{Kolmogorov-Smirnov Test}
We adopt the Kolmogorov-Smirnov (KS) test to check whether our model's input and output sequences come from the same distribution. The KS test is a nonparametric test for checking the the equality of probability distributions. In essence, the test answers the following question ``Are these two sets of samples drawn from the same distribution?''. 
The Kolmogorov-Smirnov statistic is defined as:
$$
D_{n, m}=\sup _{x}\left|F_{1, n}(x)-F_{2, m}(x)\right|, 
$$
where $\sup$ is the supremum function, $F_{1, n}$ and $F_{2, m}$ are the empirical distribution functions of the two compared samples.
For samples that are large enough, the null hypothesis would be rejected at level $\alpha$ if
$$
D_{n, m}>\sqrt{-\frac{1}{2}\ln \left(\frac{\alpha}{2}\right)} \cdot \sqrt{\frac{n+m}{n \cdot m}},
$$
where $n$ and $m$ are the first and second sample sizes.
\subsection{Distribution Analysis}
In this section, we evaluate the distribution similarity between models' input and output sequences using the KS test. In Table \ref{tab:KStest_small}, we applied the Kolmogrov-Smirnov test to check if output sequence of various models that trained on ETTm1/ETTm2 are consistent with the input sequence. 
On both datasets, by setting the standard P-value as 0.01, various existing baseline models have much smaller P-values except FEDformer and Autoformer, which indicates their outputs have a high probability of being sampled from a different distributions compared to their input signals. Autoformer and FEDformer have much larger P-values mainly due to their seasonal-trend decomposition mechanism. The proposed FiLM also has a much larger P-value compared to most baseline models. And its null hypothesis can not be rejected in most cases for these two datasets. It implies that the output sequence generated by FiLM shares a similar pattern as the input signal, and thus justifies our design motivation of FiLM as discussed in Section~\ref{sec_intro}. Though FiLM gets a smaller P-value than FEDformer, it is close to the actual output, which indicates that FiLM makes a good balance between recovering and forecasting. 

\begin{table}[h]
\centering
\caption{P-values of Kolmogrov-Smirnov test of different Transformer models for long-term forecasting output on ETTm1 and ETTm2 dataset. Larger value indicates the hypothesis (the input sequence and forecasting output come from the same distribution) is less likely to be rejected. The largest results are highlighted.}
\begin{small}
\vskip 0.05in
\scalebox{0.80}{
\begin{tabular}{c|c|cccccc}
\toprule
\multicolumn{2}{c|}{Methods}&Transformer&Informer&Autoformer&FEDformer&FiLM&True\\ 
\midrule
\multirow{4}{*}{\rotatebox{90}{ETTm1}} 
& 96 & 0.0090 & 0.0055&0.020 & \textbf{0.048} & 0.016& {0.023} \\
& 192 & 0.0052 & 0.0029&0.015 & \textbf{0.028} & 0.0123& {0.013} \\
& 336 & 0.0022 & 0.0019&0.012 & \textbf{0.015} & 0.0046& {0.010} \\
& 720 & 0.0023 & 0.0016&0.008 & \textbf{0.014} & 0.0024& {0.004} \\
\midrule
\multirow{4}{*}{\rotatebox{90}{ETTm2}} 
& 96 & 0.0012 & 0.0008& \textbf{0.079} &{0.071} & 0.022 & {0.087} \\
& 192 & 0.0011 & 0.0006& \textbf{0.047} &{0.045}  &0.020 & {0.060} \\
& 336 & 0.0005 & 0.00009& 0.027 &\textbf{0.028} &0.012 & {0.042} \\
& 720 & 0.0008& 0.0002& \textbf{0.023} & {0.021} &0.0081 & {0.023} \\
\bottomrule
\end{tabular}
}
\label{tab:KStest_small}
\end{small}
\vskip -0.1in
\end{table}

\section{Learnable Parameter Size}
\label{app:sec:learnable_para_size}

Compared to Transformer-based baseline models, FiLM enjoys a lightweight property with \textbf{80\%} learnable parameter reduction as shown in Table \ref{tab:size}. It has the potential to be used in mobile devices, or, in some situations where a lightweight model is preferred.

\begin{table}[h]
\centering
\caption{Parameter size of baseline models and FiLM with different low-rank approximations: the models are trained and tested on ETT dataset; the subscript number denotes $k$ in low-rank approximation.}
\begin{small}
\vskip 0.05in
\begin{tabular}{c|ccccccc}
\toprule
Methods&Transformer&Autoformer&FEDformer&FiLM&$\mathrm{FiLM}_{16}$ & $\mathrm{FiLM}_4$ & $\mathrm{FiLM}_1$\\
\midrule
Parameter(M) & 0.0069 & 0.0069& 0.0098& 1.50 & 0.0293 &0.0062& 0.00149 \\
\bottomrule
\end{tabular}
\label{tab:size}
\end{small}
\vskip -0.1in
\end{table}

\section{Training Speed and Memory Usage}
\label{app:time}

\begin{figure*}[h]
    \centering
    \includegraphics[width=0.45\linewidth]{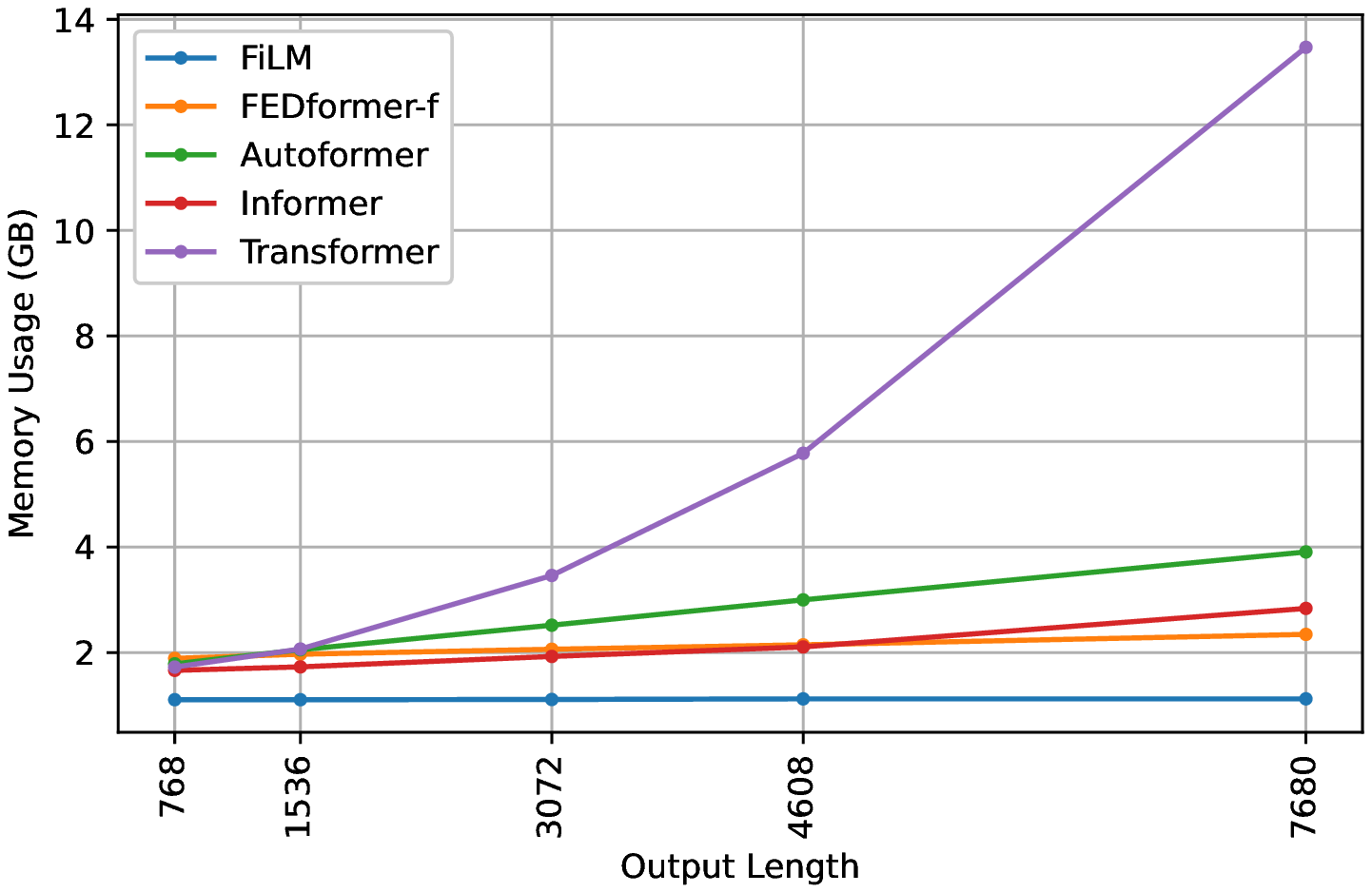}
    \includegraphics[width=0.45\linewidth]{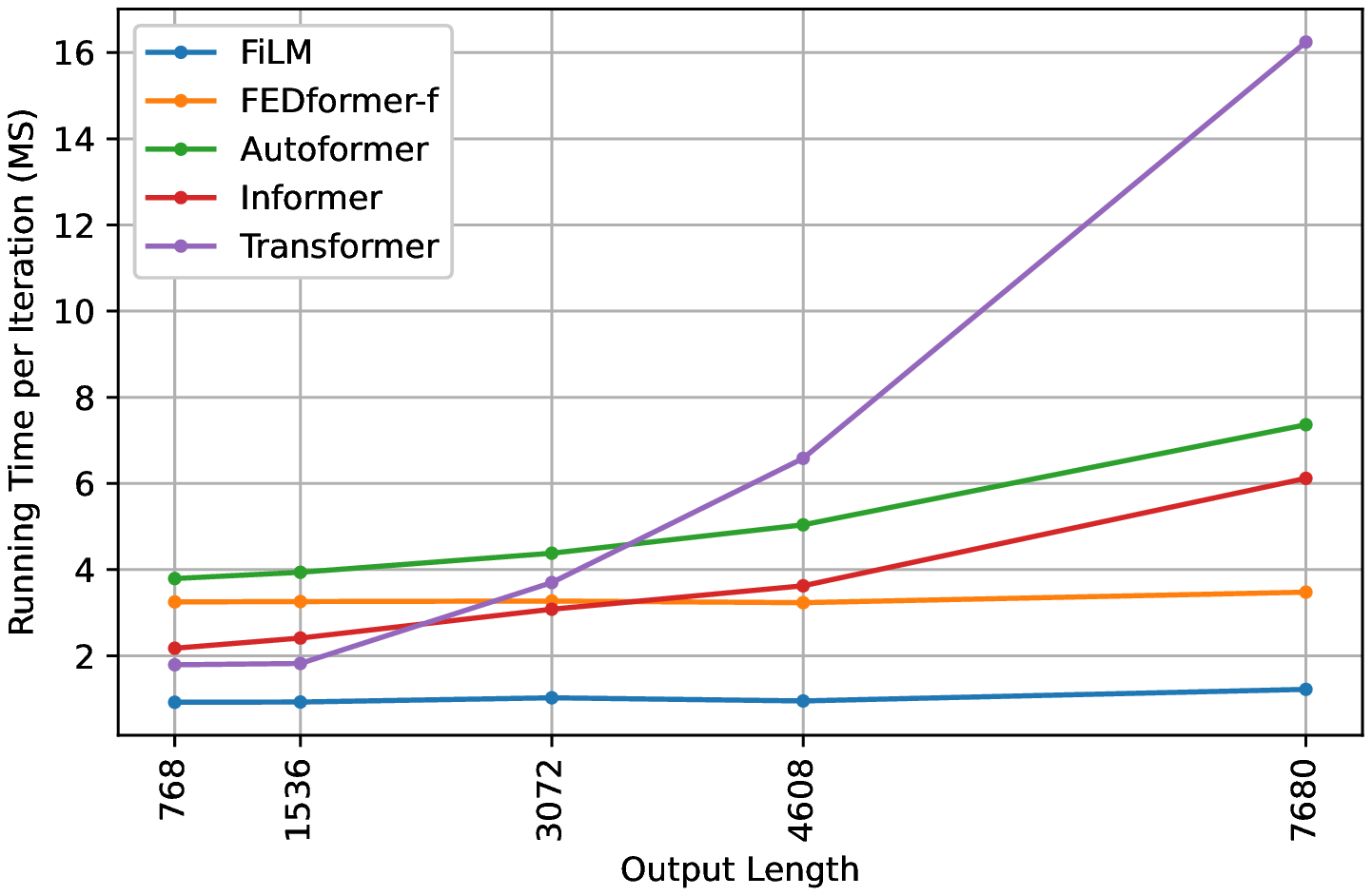}
    \caption{(Left) the memory usage of FiLM and baseline models. (Right) training speed of FiLM and baseline models. The input length is fixed to 96 and the output length is 768, 1536, 3072, 4608, and 7680.}
    \label{fig:Mem_Time}
\end{figure*}

\begin{figure*}[h]
    \centering
    \includegraphics[width=0.7\linewidth]{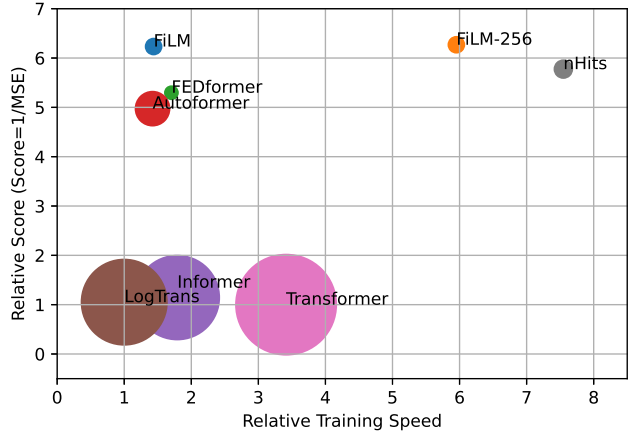}
    \caption{Comparison of training speed and performance of benchmarks. The experiment is performed on ETTm2 with output length = 96, 192, 336, and 720. The performance of the models is measured with \textit{Score}, where $Score=1/MSE$. The radius of the circle measured the STD of the performance. A higher \textit{Score} indicates better performance, same for \textit{Speed}. A smaller circle indicates better robustness. The \textit{Speed} and \textit{Score} are presented on relative value.}
    \label{fig:Time_score_scatter}
\end{figure*}

\paragraph{Memory Usage}
As shown in Figure \ref{fig:Mem_Time} (Left), FiLM has a good memory usage with the prolonging output length. For fair comparison, we fix the experimental settings of \textit{Xformer}, where we fix the input length as 96 and prolong the output length. From Figure \ref{fig:Mem_Time} (Left), we can observe that FiLM has a quasi constant memory usage. Note that the memory usage of FiLM is only linear to input length. Furthermore, FiLM enjoys a much smaller memory usage than others because of the simple architecture and compressed parameters with low-rank approximation as discussed in Appendix~\ref{app:sec:learnable_para_size}.

\paragraph{Training Speed}
Experiments are performed on one NVIDIA V100 32GB GPU. As shown in Figure \ref{fig:Mem_Time} (Right), FiLM has faster training speed than others with the prolonging output length. For fair comparison, we fix the experimental setting of \textit{Xformer}, where we fix the input length as 96 and prolong the output length. However, in the real experiment settings, we use longer input length (much longer than 96). Thus, the experiment in Figure \ref{fig:Mem_Time} (Right) is merely a toy case to show the tendency. In Figure \ref{fig:Time_score_scatter}, we show the average epoch time vs average performance under the settings of benchmarks. The experiment is performed on ETTm2 dataset with output length = 96, 192, 336, and 720. Because of the extreme low memory usage of FiLM, it can be trained with larger batch size (batch size = 256) on only one GPU compared with baselines (batch size = 32). In Figure \ref{fig:Time_score_scatter}, FiLM-256 is the FiLM models trained with batch size = 256, it exhibits significant advantages on both speed and accuracy. Furthermore, due to the shallow structure and smaller amount of trainable parameters, FiLM is easy to converge and enjoys smaller performance variation and smaller performance degradation when using large batch size. It is observed that the models with Fourier enhanced block (FiLM \& FEDformer) have better robustness. It also worth noting that vanilla Transformer has good training speed because of the not-so-long sequence length. Only a sequence length over one thousand will distinguish the advantage of efficient Transformers.



\section{Additional Benchmarks}
\subsection{Multivariate long-term series forecasting with extra baseline models}
For the additional benchmarks for multivariate experiments, we add some non-Transformer methods for comparison.N-BEATS\cite{nbeats} and N-HiTS\cite{challu2022n} are two recent proposed powerful non-Transformer methods. As N-HiTS is the latest development from the research group, which also published N-BEATS, we add N-HiTS to our empirical comparison. Here, we adopt the results in the N-HiTS paper to prevent inappropriate parameter tuning problems. We also add a seasonal-naive model in the comparison. FiLM outperforms N-HiTS in most cases(33/48). Moreover, Simple Seasonal-naive\cite{seasonal-naive} is a solid baseline on exchange datasets better than N-hits, Fedformer, and Autoformer, but FiLM still surpasses its performance, as shown in Table \ref{tab:multi-benchmarks}.
\begin{table*}[h!]
\centering
\caption{multivariate long-term series forecasting results on six datasets with various input length and prediction length $O \in \{96,192,336,720\}$ (For ILI dataset, we set prediction length $O \in \{24,36,48,60\}$). Supplementary results of non-Transformer baselines (N-Hits and a seasonal-naive model).}
\scalebox{0.8}{
\begin{tabular}{c|c|cccccccccccccccc|}
\toprule
\multicolumn{2}{c|}{Methods}&\multicolumn{2}{c|}{FiLM}&\multicolumn{2}{c|}{N-Hits}&\multicolumn{2}{c|}{FEDformer}&\multicolumn{2}{c|}{Autoformer}&\multicolumn{2}{c|}{Seasonal-naive}\\
\midrule
\multicolumn{2}{c|}{Metric} & MSE  & MAE & MSE & MAE& MSE  & MAE& MSE  & MAE& MSE  & MAE \\
\midrule
\multirow{4}{*}{\rotatebox{90}{$ETTm2$}} 
                        &96 & \textbf{0.165} & 0.256 & 0.176 & \textbf{0.255} &0.203 &0.287 &0.255  &0.339  &0.262 &0.300 \\
                        & 192 & \textbf{0.222} & \textbf{0.296} & 0.245 & 0.305 & 0.269  &0.328  &0.281 &0.340  &0.319 &0.337\\
                        & 336 & \textbf{0.277} & \textbf{0.333} & 0.295 & 0.346 & 0.325 &0.366 &0.339  &0.372  &0.375 &0.371\\
                        & 720 & \textbf{0.371} & \textbf{0.389} & 0.401 & 0.426 & 0.421 &0.415 &0.422  &0.419 &0.469 &0.422\\
\midrule
\multirow{4}{*}{\rotatebox{90}{$Electricity$}} 
                        &96  &0.154  &0.267 & \textbf{0.147} & \textbf{0.249}  &0.183 &0.297 &0.201  &0.317 &0.211 &0.278\\
                        & 192 &\textbf{0.164} & \textbf{0.258} & 0.167 & 0.269 & 0.195 &0.308 &0.222  &0.334 &0.214 &0.282\\
                        & 336 & 0.188 & \textbf{0.283} & \textbf{0.186} & 0.290 & 0.212 &0.313 &0.231 &0.338 &0.226 &0.294\\
                        & 720 & 0.236 & \textbf{0.332} & 0.243 & 0.340 &\textbf{0.231} &0.343 &0.254  &0.361 &0.265 & 0.324\\
\midrule
\multirow{4}{*}{\rotatebox{90}{$Exchange$}} 
                        &96  & \textbf{0.086} & \textbf{0.204} &0.092 & 0.211 & 0.139 & 0.276 & 0.197 &0.323 &0.086 &0.204 \\
                        & 192 & \textbf{0.189} & \textbf{0.292} & 0.208 & 0.322 &0.256 &0.369 & 0.300 & 0.369 &0.172 &0.295\\
                        & 336 & \textbf{0.356} & \textbf{0.433} & 0.371 & 0.443 & 0.426 &0.464 & 0.509 & 0.524 &0.311 &0.401\\
                        & 720 & \textbf{0.727} & \textbf{0.669} & 0.888 & 0.723 &  1.090 &0.800 & 1.447 & 0.941 &0.832 &0.686\\
\midrule
\multirow{4}{*}{\rotatebox{90}{$Traffic$}} 
                        &96  & 0.416& 0.294 & \textbf{0.402} & \textbf{0.282} & 0.562 &0.349 &0.613  &0.388 &1.219 &0.497\\
                        & 192 & \textbf{0.408} & \textbf{0.288} &0.420 & 0.297 & 0.562 &0.346 &0.616&0.382 &1.089 &0.456\\
                        & 336 & \textbf{0.425} & \textbf{0.298} &0.448 & 0.313 & 0.570 &0.323 &0.622  &0.337 &1.147 &0.473\\
                        & 720 & \textbf{0.520} &\textbf{0.353} &0.539 & 0.353 & 0.596 &0.368 &0.660  &0.408 &1.181 &0.486\\
\midrule
\multirow{4}{*}{\rotatebox{90}{$Weather$}} 
                        & 96 & 0.199 & 0.262 & \textbf{0.158} & \textbf{0.195} & 0.217  &0.296  &0.266  &0.336 &0.315 &0.288\\
                        & 192 & 0.228 & 0.288 & \textbf{0.211} & \textbf{0.247} & 0.276  &0.336  &0.307  &0.367 &0.341 &0.305\\
                        & 336 &\textbf{0.267} & 0.323 & 0.274 & \textbf{0.300} & 0.339  &0.380  &0.359  &0.395 &0.381 &0.331\\
                        & 720 &\textbf{0.319} & 0.361 &0.351 & \textbf{0.353} & 0.403  &0.428 &0.578 &0.578 &0.440 &0.370 \\
\midrule
\multirow{4}{*}{\rotatebox{90}{$ILI$}} 
                        & 24 & 1.970 & 0.875 & \textbf{1.862} & \textbf{0.869} & 2.203  &0.963  &3.483 &1.287 &6.581 &1.699 \\
                        & 36 & \textbf{1.982} & \textbf{0.859} &2.071 & 0.969 & 2.272  &0.976  &3.103  &1.148 &7.121 &1.882 \\
                        & 48 & \textbf{1.868} & \textbf{0.896} &2.346 & 1.042 & 2.209  &0.981  &2.669  &1.085 &6.567 &1.797 \\
                        & 60 & \textbf{2.057} & \textbf{0.929} &2.560 & 1.073 & 2.545  &1.061  &2.770  &1.125 &5.885 &1.675 \\
\bottomrule
\end{tabular}
\label{tab:multi-benchmarks}
}
\end{table*}
\subsection{Ablation univariate forecasting experiments for FEL layers with all six datasets }
As shown in Table \ref{tab:ablation_FEL_full}, although LPU+MLP combining all boosting tricks has slightly better performance than FiLM for the ETTm1 dataset, FiLM remains the most consistent and effective model among all variants across all six datasets. FEL is a much better backbone structure than MLP, LSTM, CNN, and vanilla attention modules. 

\begin{table*}[!h]
\vskip -0.05in
\centering
\caption{: (Full Benchmark)Ablation studies of FEL layer. The FEL layer is replaced with 4 different variants: MLP, LSTM, CNN, and Transformer.The experiments are performed on ETTm1 and Electricity. The metric of variants is presented in relative value (’+’ indicates degraded performance, ’-’ indicates improved performance).}\vspace{-1mm}
\scalebox{0.8}{
\begin{tabular}{c|c|ccccccccccccccc}
\toprule
\multicolumn{2}{c|}{Methods}&\multicolumn{2}{c|}{FilM}&\multicolumn{2}{c|}{LPU+MLP}&\multicolumn{2}{c|}{LPU+LSTM}&\multicolumn{2}{c|}{LPU+CNN}&\multicolumn{2}{c|}{LPU+attention}\\
\midrule
\multicolumn{2}{c|}{Metric} & MSE  & MAE & MSE & MAE& MSE  & MAE& MSE  & MAE& MSE  & MAE& \\
\midrule

\multirow{4}{*}{\rotatebox{90}{ETTm1}} 

&96  &\textbf{0.029} &\textbf{0.127}  &+0.0\%   & +0.0\%   & +12.1\% & +7.1\%    & +13.5\%   & +9.5\%   & +1.7\%  & +1.6\%  \\
& 192  &\textbf{0.041} &\textbf{0.153}  &-1.5\% & -0.6\%  & +12.2\% & +8.5\%  & +10.8\%   & +7.8\%    & +2.0\%  & +3.3\%  \\
& 336 & \textbf{0.053} & \textbf{0.175} &-1.7\% & -1.7\%  & +4.5\% & +4.0\%    & +10.8\%    & +6.3\%    & +4.5\% & +2.9\%   \\
& 720 &\textbf{0.071}& \textbf{0.205}   &-0.9\%  & -1.0\%  & +5.5\% & +3.4\%  & +8.6\%    & +4.9\%    & +13.8\% & +7.8\%   \\

\midrule
\multirow{4}{*}{\rotatebox{90}{Electricity}} 

&96  &\textbf{0.154}  &\textbf{0.247}  &+155\%  & +81\%   & +160\% & +84\%    & +330\%  & +155\%  & +242\% & +119\%  \\
& 192 &\textbf{0.166} & \textbf{0.258} &+59\% & +39\%   & +121\%  & +67\%   & +224\%  & +117\%  & +264\% & +131\%  \\
& 336 & \textbf{0.188}& \textbf{0.283} &+55\%  & +35\%   & +150\%  & +74\%   & +128\%  & +71\% & +183\% & +95\%  \\
& 720 & \textbf{0.249}&\textbf{0.341}  &+33\%  & +25\%   & +154\%  & +73\%   & +192\%  & +95\%  & +312\% & +138\%  \\ 

\midrule
\multirow{4}{*}{\rotatebox{90}{Exchange}} 

&96  &\textbf{0.110}  &\textbf{0.247}  &-13\%  & -12\%   & +51\% & +17\%    & +4.6\%  & -1.2\%  & -4.6\% & -5.8\%  \\
& 192 &\textbf{0.207} & \textbf{0.352} &+7.2\% & +0.0\%   & +69\%  & +32\%   & +29\%  & +12\%  & +22\% & +11\%  \\
& 336 & \textbf{0.327}& \textbf{0.461} &+48\%  & +13\%   & +62\%  & +20\%   & +68\%  & +24\% & +72\% & +23\%  \\
& 720 & \textbf{0.811}&\textbf{0.708}  &+29\%  & +14\%   & +24\%  & +9.6\%   & +38\%  & +12\%  & +64\% & +27\%  \\ 

\midrule
\multirow{4}{*}{\rotatebox{90}{Traffic}} 

&96  &\textbf{0.144}  &\textbf{0.215}  &+69\%  & +47\%   & +13\% & +15\%    & +300\%  & +176\%  & +271\% & +161\%  \\
& 192 &\textbf{0.120} & \textbf{0.199} &+17\% & +7.5\%   & +31\%  & +24\%   & +258\%  & +149\%  & +1572\% & +355\%  \\
& 336 & \textbf{0.128}& \textbf{0.212} &+6.2\%  & +7.6\%   & +16\%  & +15\%   & +151\%  & +102\% & +1514\% & +368\%  \\
& 720 & \textbf{0.153}&\textbf{0.252}  &+38\%  & +28\%   & +11\%  & +7.9\%   & +250\%  & +126\%  & +1048\% & +349\%  \\ 

\midrule
\multirow{4}{*}{\rotatebox{90}{Weather}} 

&96  &\textbf{0.0012}  &\textbf{0.026}  &+17\%  & +6.2\%   & +16\% & +6.9\%    & +19\%  & +8.1\%  & +21\% & +8.9\%  \\
& 192 &\textbf{0.0014} & \textbf{0.029} &-1.4\% & -2.4\%   & +5.0\%  & +1.7\%   & 0.7\%  & -0.7\%  & +4.3\% & +1.4\%  \\
& 336 & \textbf{0.0015}& \textbf{0.030} &+0.0\%  & -0.6\%   & +3.3\%  & +1.3\%   & +2.0\%  & +0.0\% & +3.3\% & +1.3\%  \\
& 720 & \textbf{0.0022}&\textbf{0.037}  &+4.6\%  & -0.3\%   & +4.1\%  & -1.6\%   & 3.6\%  & 0.0\%  & +0.0\% & -3.8\%  \\ 

\midrule
\multirow{4}{*}{\rotatebox{90}{ILI}} 

&96  &\textbf{0.629}  &\textbf{0.538}  &+51\%  & +45\%   & -2.5\% & 9.5\%    & +20\%  & +29\%  & +112\% & +59\%  \\
& 192 &\textbf{0.444} & \textbf{0.481} &+99\% & +58\%   & +25\%  & +24\%   & +84\%  & +56\%  & +360\% & +142\%  \\
& 336 & \textbf{0.557}& \textbf{0.584} &+33\%  & +31\%   & +21\%  & +16\%   & +58\%  & +30\% & +702\% & +94\%  \\
& 720 & \textbf{0.641}&\textbf{0.644}  &+8.4\%  & +5.4\%   & +23\%  & +18\%   & +42\%  & +22\%  & +74\% & +34\%  \\ 
\bottomrule
\end{tabular}
\label{tab:ablation_FEL_full}
}

\vskip -0.05in
\end{table*}

\subsection{Boosting experiments of LPU with common deep learning backbones for all six datasets}
As shown in Table \ref{tab:boosting}, LPU shows a consistent boosting effect across all selected common deep learning backbones for most datasets. It can be used as a simple and effective build add-on block for long-term time series forecasting tasks. Although without data normalization, pure LPU negatively boosts performance for some cases. 

\vskip -0.1in

\begin{table*}[h]
\centering
\caption{(Full Benchmark) Boosting effect of LPU layer for common deep learning backbones: MLP, LSTM, CNN and Attention.`+` indicates degraded performance.}\vspace{-1mm}
\scalebox{0.80}{
\begin{tabular}{c|c|cccccccccccccccccc}
\toprule
\multicolumn{2}{c|}{Methods}&\multicolumn{2}{c|}{FEL}&\multicolumn{2}{c|}{MLP}&\multicolumn{2}{c|}{LSTM}&\multicolumn{2}{c|}{lagged-LSTM}&\multicolumn{2}{c|}{CNN}&\multicolumn{2}{c}{Attention}\\
\midrule
\multicolumn{2}{c|}{Compare} & LPU  & Linear& LPU  & Linear & LPU  & Linear & LPU  & Linear & LPU  & Linear & LPU  & Linear\\
\midrule

\multirow{4}{*}{\rotatebox{90}{ETTm1}} 

&96  &\textbf{0.030}  & +38\%   & 0.034  & +8.0\%  & 0.049 & +73\% & 0.093 & -21\% & 0.116   & -50\%   & 0.243  & -81\% \\
& 192  &\textbf{0.047} & +9.5\% & 0.049  & +30\%  & 0.174 & +32\% & 0.331 & -48\% & 0.101   & +20\%    & 0.387  & -86\% \\
& 336 & 0.063 & +5.8\% & \textbf{0.061}  & +64\%  & 0.119 & +84\% & 0.214 & -19\% & 0.122   & +25\%    & 1.652  & +12\% \\
& 720 &\textbf{0.081} & +1.4\%  & 0.082  & +62\%  & 0.184 & +32\% & 0.303 & -6.5\%  & 0.108   & +13\%    & 4.782  & -61\% \\

\midrule
\multirow{4}{*}{\rotatebox{90}{Electricity}} 

&96  &\textbf{0.213}  & +136\%  & 0.431  & +121\%   & 0.291 & +56\% & 0.739 & -33\% & 0.310  & +43\%  & 0.805 & +23\% \\
& 192 &\textbf{0.268} & +32\%   & 0.291 & +239\%   & 0.353  & +17\% & 0.535 & +15\% & 0.380  & +12\%  & 0.938 & +14\% \\
& 336 & 0.307& +0.1\%  & \textbf{0.296} & +235\%   & 0.436  & -6.7\% & 0.517 & +23\% & 0.359  & +29\% & 2.043 & -54\% & \\
& 720 & \textbf{0.321}& +37\%   & 0.339 & +196\%   & 0.636  & -11\% & 0.492 & +28\% & 0.424  & +18\%  & 9.115 & +298\% \\ 
               
\midrule
\multirow{4}{*}{\rotatebox{90}{$Exchange$}} 
& 96  & 0.130 & +7.5\% & \textbf{0.110} & -18\% & 0.224 & +6.0\% & 0.521 & -58\% & 0.244 & -18\% & 0.338 & +872\% \\
& 192 & \textbf{0.205} & +39\% & 0.257 & -36\% & 0.787 & -35\% & 1.742 & -66\% & 0.630 & +2.1\% & 0.930 & +278\% \\
& 336 & 0.467 & +9.2\% & \textbf{0.461} & -33\% & 0.964 & +24\% & 2.281 & -38\% & 3.231 & -85\% & 1.067 & +69\%\\
& 720 & \textbf{1.003} & +26\% & 1.981 & -61\% & 2.703 & -29\% & 1.457 & +34\% & 5.531 & +9.7\% & 0.631 & +1831\% \\
\midrule

\multirow{4}{*}{\rotatebox{90}{$Traffic$}} 
&96  &\textbf{0.312} & +18\% & 0.376 & +277\% & 0.215 & +1.2\% & 0.216 & +10\% & 0.543 & -33\% & 0.429 & +210\% \\
& 192 & \textbf{0.141} & +9.6\% & 0.199 & +598\% & 0.177 & +19\% & 0.186 & +17\% & 0.451 & +9.0\% & 0.476 & +176\% \\
& 336 & \textbf{0.143} & +2.5\% & 0.195 & +613\% & 0.192 & +19\% & 0.190 & +11\% & 0.346 & +44\% & 0.377 & +260\% \\
& 720 & \textbf{0.215} & +30\% & 0.240 & +475\% & 0.234 & -1.7\% & 0.250 & +15\% & 0.348 &  +47\% & 0.773 & +171\%\\
\midrule

\multirow{4}{*}{\rotatebox{90}{$Weather$}} 
& 96 & 0.0073 & -38\% & 0.006  & -33\% & 0.006 & -23\% & 0.0070 & -17\% & \textbf{0.0022} & +167\% & 0.0065 &-11\%\\
& 192 & 0.0106 & -64\% & 0.007 & -14\% & 0.0074 & -11\% & \textbf{0.0063} & -19\% & 0.007 & -24\% & 0.0075 & -12\%\\
& 336 & 0.0079 & -37\% & 0.006 & +4.9\% & 0.0056 & +12\% & \textbf{0.0055} & +12\% & 0.0056 & +0.5\% & 0.222 & -69\% \\
& 720 &0.0063 & +0.4\% & 0.006 & +7.6\% & \textbf{0.0062} & +5.3\% & 0.103 & -36\% &0.006 & +4.2\% & 0.037 & -81\% \\
                        
\midrule
\multirow{4}{*}{\rotatebox{90}{$ILI$}} 
& 24 & 1.393 &	+6.1\% & \textbf{1.220} & +36\% & 2.306 & +66\% & 4.189 & -9.2\% & 2.264 & -22\% & 2.249 & +217\% \\
& 36 & 1.242 & -22\% & \textbf{1.185} & +56\% & 2.950 & +44\% & 2.516 & +42\% & 1.841 & -3.0\% & 5.026 & +45\% \\
& 48 & 1.448 & -28\% & \textbf{1.079} & +79\% & 3.385 & +38\% & 3.501 & +16\% & 1.654 & +23\% & 2.838 & +115\% \\
& 60 & 2.089 & -18\% & \textbf{0.986} & +96\%  & 4.031 & +18\% & 4.258 & +10\% & 1.290 & +176\% & 4.978 & +250\% \\
\bottomrule
\end{tabular}
\label{tab:boosting}
}
\end{table*}

\subsection{Ablation univariate forecasting experiments for Low rank approximation with all six datasets}
As shown in Table \ref{tab:FEL_LRA_FULL}, with the low-rank approximation of learnable matrix in Fourier Enhanced Layer significantly reducing our parameter size, and even improve our model's performance for in some datasets.
\begin{table*}[h]

\centering
\caption{Low-rank Approximation (LRA) univariate forecasting study for frequency enhanced layer: Comp. K=0 means default version without LRA, 1 means the largest compression using K=1.}
\scalebox{0.75}{
\begin{tabular}{c|c|cccccccccc}
\toprule

\multicolumn{2}{c|}{Comp. K}&\multicolumn{2}{c|}{0}&\multicolumn{2}{c|}{16}&\multicolumn{2}{c|}{4}&\multicolumn{2}{c}{1}\\

\midrule
\multicolumn{2}{c|}{Metric} & MSE  & MAE & MSE & MAE& MSE  & MAE& MSE  & MAE \\
\midrule


\multirow{4}{*}{\rotatebox{90}{$ETTm2$}} 

&96&0.065&0.189&0.064&0.185&\textbf{0.064}&\textbf{0.185}&0.064&0.186 \\

&192&0.094&0.233&0.094&0.231&0.093&0.231&\textbf{0.093}&\textbf{0.231} \\

&336&0.124&0.274&0.124&0.270&\textbf{0.124}&\textbf{0.269}&0.124&0.271 \\

&720&0.173&0.323&0.173&0.322&\textbf{0.173}&\textbf{0.322}&0.177&0.328 \\
\midrule

\multirow{4}{*}{\rotatebox{90}{$Electricity$}} 

&96&\textbf{0.154}&\textbf{0.247}&0.211&0.324&0.216&0.331&0.277&0.387 \\

&192&\textbf{0.166}&\textbf{0.258}&0.251&0.352&0.246&0.347&0.334&0.421 \\

&336&\textbf{0.188}&\textbf{0.283}&0.276&0.369&0.302&0.396&0.363&0.440 \\

&720&\textbf{0.249}&\textbf{0.341}&0.336&0.429&0.342&0.436&0.411&0.481 \\

\midrule

\multirow{4}{*}{\rotatebox{90}{$Exchange$}} 

&96&0.110&0.259&0.119&0.273&\textbf{0.104}&\textbf{0.247}&0.105&0.251 \\

&192&0.207&0.352&0.196&0.355&\textbf{0.195}&\textbf{0.349}&0.212&0.372 \\

&336&\textbf{0.327}&\textbf{0.461}&0.388&0.497&0.373&0.491&0.407&0.506 \\

&720&\textbf{0.811}&\textbf{0.708}&0.908&0.767&1.288&0.941&1.840&1.153 \\
\midrule

\multirow{4}{*}{\rotatebox{90}{$Traffic$}} 

&96&\textbf{0.144}&\textbf{0.215}&0.146&0.223&0.154&0.237&0.267&0.373 \\

&192&\textbf{0.120}&\textbf{0.199}&0.121&0.201&0.138&0.231&0.218&0.333 \\

&336&0.128&0.212&\textbf{0.120}&\textbf{0.206}&0.132&0.227&0.216&0.335 \\

&720&\textbf{0.153}&\textbf{0.252}&0.155&0.257&0.154&0.257&0.246&0.366 \\

\midrule

\multirow{4}{*}{\rotatebox{90}{$Weather$}} 

&96&0.0012&0.026&0.0011&0.025&\textbf{0.001}&\textbf{0.025}&0.001&0.025 \\

&192&0.0014&0.029&0.0014&0.028&\textbf{0.001}&\textbf{0.028}&0.001&0.028 \\

&336&0.0015&0.03&0.0015&0.030&\textbf{0.001}&\textbf{0.029}&0.002&0.030 \\

&720&0.0022&0.037&0.0022&0.037&\textbf{0.002}&\textbf{0.037}&0.002&0.037 \\
\midrule

\multirow{4}{*}{\rotatebox{90}{$ILI$}} 

&96&0.629&0.538&\textbf{0.599}&\textbf{0.556}&0.628&0.558&0.630&0.579 \\

&192&\textbf{0.444}&\textbf{0.481}&0.487&0.533&0.508&0.561&0.570&0.612 \\

&336&0.557&0.584&\textbf{0.553}&\textbf{0.565}&0.703&0.696&0.722&0.706 \\

&720&\textbf{0.641}&\textbf{0.644}&0.648&0.641&0.900&0.780&1.493&1.032 \\

\midrule
\multicolumn{2}{c|}{Parameter size}&\multicolumn{2}{c|}{100\%}&\multicolumn{2}{c|}{6.4\%}&\multicolumn{2}{c|}{1.6\%}&\multicolumn{2}{c}{0.4\%}\\
\bottomrule
\end{tabular}
\label{tab:FEL_LRA_FULL}
}
\end{table*}

\subsection{Ablation univariate forecasting experiments for frequency mode selection with all six datasets}
Three different mode selection policies are studied for frequency enhanced layer: 1) lowest mode selection: we select $m$ lowest frequency modes to retain. 2) random model selection: we select $m$ frequency modes randomly to retain. 3) lowest with extra high mode selection: we select $0.8\times m$ lowest frequency modes and $0.2\times m$ high frequency modes randomly to retain. The experimental results are summarized in Table \ref{tab:mode_full} with $m = 64$ for both experiments. Lowest mode selection is the most stable frequency mode selection policy through adding some randomness mode can improve the results for some datasets.


\begin{wraptable}[11]{r}{0.5\textwidth}
\vskip -0.1in

\centering
\caption{Mode selection policy study for frequency enhanced layer. Lowest: select the lowest $m$ frequency mode; Random: select $m$ random frequency mode; Low random: select the $0.8*m$ lowest frequency mode and $0.2*m$ random high frequency mode.}\vspace{-1.5mm}
\scalebox{0.65}{
\begin{tabular}{c|c|cccccccc}
\toprule
\multicolumn{2}{c|}{Policy}&\multicolumn{2}{c|}{Lowest}&\multicolumn{2}{c|}{Random}&\multicolumn{2}{c}{Low random}\\
\midrule
\multicolumn{2}{c|}{Metric} & MSE  & MAE & MSE & MAE& MSE  & MAE \\
\midrule
\multirow{4}{*}{\rotatebox{90}{ETTm2}} 
&96&\textbf{0.065}&\textbf{0.189}&0.066&0.189&0.066&0.190 \\

&192&\textbf{0.094}&\textbf{0.233}&0.096&0.235&0.096&0.235 \\

&336&\textbf{0.124}&0.274&0.125&\textbf{0.270}&0.128&0.275 \\

&720&0.173&0.323&\textbf{0.173}&\textbf{0.322}&0.173&0.323 \\

\midrule
\multirow{4}{*}{\rotatebox{90}{Electricity}} 
&96&\textbf{0.154}&\textbf{0.247}&0.175&0.0.260&0.176&0.262 \\

&192&\textbf{0.166}&\textbf{0.258}&0.177&0.266&0.168&0.273 \\

&336&\textbf{0.188}&\textbf{0.283}&0.199&0.289&0.192&0.299 \\

&720&\textbf{0.249}&\textbf{0.341}&0.269&0.364&0.270&0.362 \\

\midrule
\multirow{4}{*}{\rotatebox{90}{Exchange}} 
&96&0.11&0.259&0.110&0.256&\textbf{0.106}&\textbf{0.249} \\

&192&0.207&0.352&\textbf{0.196}&\textbf{0.351}&0.207&0.357 \\

&336&\textbf{0.327}&\textbf{0.461}&0.451&0.522&0.373&0.484 \\

&720&0.811&0.708&0.835&0.714&\textbf{0.604}&\textbf{0.628} \\

\midrule
\multirow{4}{*}{\rotatebox{90}{Traffic}} 
&96&\textbf{0.144}&\textbf{0.215}&0.145&0.216&0.145&0.217 \\

&192&0.12&0.199&0.119&0.198&\textbf{0.118}&\textbf{0.197} \\

&336&0.128&0.212&\textbf{0.122}&\textbf{0.207}&0.122&0.209 \\

&720&0.153&0.252&\textbf{0.142}&\textbf{0.238}&0.155&0.259 \\

\midrule
\multirow{4}{*}{\rotatebox{90}{Weather}} 
&96&\textbf{0.0012}&\textbf{0.026}&0.0012&0.027&0.0012&0.026 \\

&192&\textbf{0.0014}&\textbf{0.029}&0.0014&0.029&0.0014&0.029 \\

&336&\textbf{0.0015}&\textbf{0.03}&0.0015&0.030&0.0015&0.030 \\

&720&\textbf{0.0022}&\textbf{0.037}&0.0023&0.037&0.0023&0.037 \\

\midrule
\multirow{4}{*}{\rotatebox{90}{ILI}} 
&96&0.629&0.538&0.639&0.542&\textbf{0.626}&\textbf{0.537} \\

&192&\textbf{0.444}&\textbf{0.481}&0.448&0.490&0.447&0.494 \\

&336&\textbf{0.557}&\textbf{0.584}&0.560&0.590&0.557&0.587 \\

&720&\textbf{0.641}&\textbf{0.644}&0.641&0.647&0.643&0.650 \\

\bottomrule

\end{tabular}
\label{tab:mode_full}
}
\end{wraptable}

\end{document}